%% file: neurips_2026_camera_ready.tex
\documentclass{article}
\usepackage{makecell}
    \PassOptionsToPackage{numbers, compress}{natbib}

 \usepackage[eandd, final]{neurips_2026}

\usepackage[utf8]{inputenc} 
\usepackage[T1]{fontenc}    
\usepackage{hyperref}       
\usepackage{url}            
\usepackage{booktabs}       
\usepackage{amsfonts}       
\usepackage{nicefrac}       
\usepackage{microtype}      
\usepackage{xcolor}         

\usepackage{bbm}
\usepackage{xspace}
\usepackage{dsfont}
\usepackage{multirow}
\usepackage{graphicx}
\usepackage{tabularx}  
\usepackage{booktabs}
\usepackage{caption}
\usepackage{subcaption} 
\usepackage{amsmath}
\usepackage{pifont}
\usepackage{tikz}
\usepackage{adjustbox}
\usepackage{wrapfig}
\usetikzlibrary{calc}

\usepackage{xurl}
\hypersetup{
    colorlinks=true,
    linkcolor=violet,
    citecolor=teal,
    filecolor=DarkSlateGray,      
    urlcolor=violet,
    pdftitle={NavAble},
    pdfpagemode=FullScreen,
}

\usepackage{navable_logo}

\hypersetup{pdftitle={NavAble: A Large-Scale Dataset and Synthetic Data Generation Pipeline for Blind Navigation}}
\title{\texorpdfstring{\NavAbleLogo}{NavAble}: A Large-Scale Dataset and Synthetic Data Generation Pipeline for Blind Navigation}

\author{%
  \textbf{%
    Hochul Hwang\textsuperscript{1}\thanks{Equal contribution.}
    \quad Jahir Sadik Monon\textsuperscript{1}\footnotemark[1]
    \quad Soowan Yang\textsuperscript{2}
    \quad Shiven Umeshbhai Patel\textsuperscript{1}
  }\\
  \textbf{%
    Anh Nhat Hong Nguyen\textsuperscript{1}
    \quad Kien Nguyen\textsuperscript{1}
    \quad Keshav Garg\textsuperscript{1}
    \quad Dylan Gage\textsuperscript{1}
  }\\
  \textbf{%
    Duretti Hordofaa\textsuperscript{1}
    \quad Anshu Anjna\textsuperscript{1}
    \quad Eshed Ohn-Bar\textsuperscript{3}
    \quad Donghyun Kim\textsuperscript{1}
  }\\
  \textsuperscript{1}University of Massachusetts Amherst\\
  \textsuperscript{2}Daegu Gyeongbuk Institute of Science and Technology\\
  \textsuperscript{3}Boston University\\
  \texttt{hochulhwang@umass.edu}
}

\begin{document}

\maketitle

\begin{abstract}

Reliable recognition of accessibility-critical objects (e.g., audible pedestrian signals, door-activation buttons, and handrails) is essential for assistive navigation technologies that support safe, independent mobility for blind and low-vision (BLV) users. Yet these categories are severely underrepresented in existing large-scale vision datasets, and the few datasets that include them suffer from limited class coverage, inconsistent annotations, and poor diversity, limiting the perception reliability that BLV navigation requires.
We introduce \textsc{NavAble}, a large-scale dataset spanning 11 accessibility-critical object classes, combining 36K curated open-source images with 5.5K newly collected, densely annotated real-world images. To extend scale and diversity, we develop a synthetic data generation pipeline that renders class-validated 3D assets across 37 environments under diverse styles and viewpoints, yielding over 450K frames with rich ground-truth annotations from 500 assets. The pipeline minimizes manual effort by combining automated web image crawling, VLM-based filtering, SAM~3D-based asset generation, and convenient camera trajectory configuration, enabling large-scale data generation at a diversity unattainable through manual collection. Experiments across seven segmentation architectures show that fine-tuning on the \textsc{NavAble} Dataset achieves up to 71.70\% foreground mIoU on our held-out real-world test set, substantially outperforming state-of-the-art open-vocabulary zero-shot models (at most 26.07\% foreground mIoU). Furthermore, adding synthetic data to real-only training improves five of the seven evaluated architectures by up to +7.19 foreground mIoU. Beyond benchmark evaluation, out-of-distribution validation on egocentric data from two mobility-assistive robots demonstrates relative transfer improvements.
Together, \textsc{NavAble} establishes a scalable foundation for BLV navigation perception, directly addressing the data scarcity and embodiment gaps that have limited progress in this domain. The dataset and pipeline are publicly available at \url{https://guidedogrobot-navable.github.io}.
\end{abstract}

\begin{figure}[t]
\centering
\includegraphics[width=\linewidth]{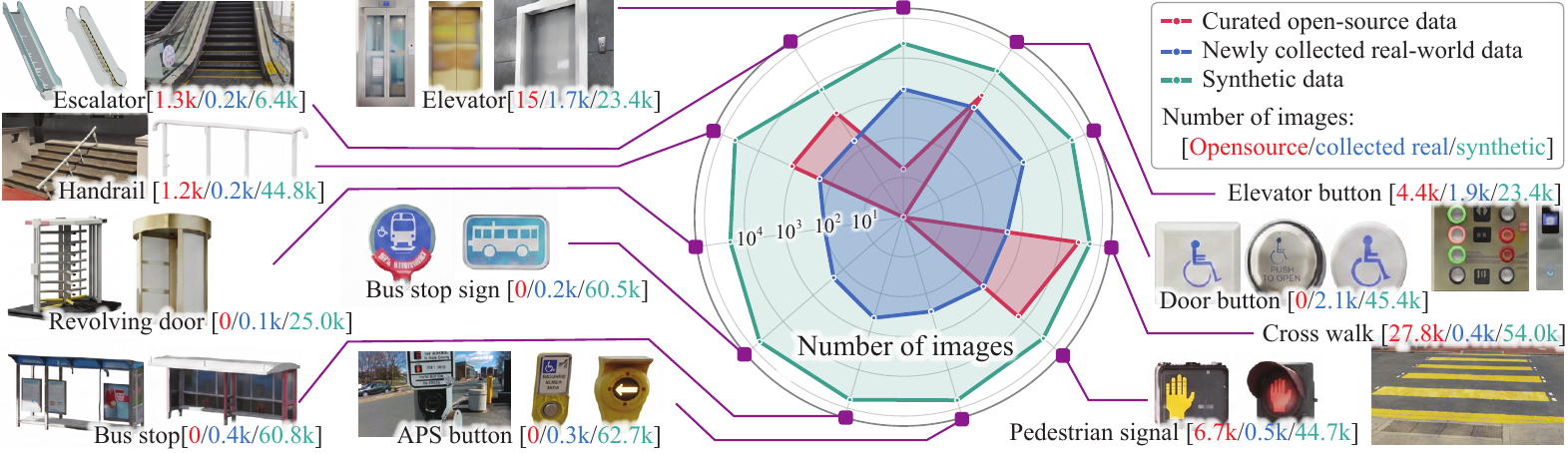}
\caption{\textbf{Overview of \textsc{NavAble} Dataset.} Existing datasets limited object classes (e.g., lacking APS or door buttons). \textsc{NavAble} Dataset provides a curated multi-source collection of open-source, newly collected real-world data, and synthetic data across 11 accessibility-critical object categories.}
\label{fig:overview}
\end{figure}

\section{Introduction} 

Over 36 million people worldwide are blind, a number projected to more than double in the next 30 years \cite{ackland2017world}. In response, researchers have developed a wide range of assistive technologies, including smart white canes~\citep{kinectcane,slade2021multimodal}, wearable computer-vision systems \citep{wearableBLV}, and guide dog robot platforms~\cite{cai2024navigating,guerreiro2019cabot,guidedog-dev}, all of which rely on robust perception to alert users to hazards, identify navigational landmarks, and support route-following. In practice, however, prior work with BLV end-users shows that perception failures remain a major barrier to adoption, as missed detections and false positives interrupt navigation and erode user trust~\cite{giudice2008blind, chi24_robotic_companions, streetnav, blv-survey}. 

At the core of this problem is a data bottleneck: the accessibility-critical object categories that matter most for BLV navigation---such as Accessible Pedestrian Signals (APS), handrails, and door-activation buttons---are either entirely absent, coarsely defined, or severely underrepresented in large-scale vision datasets such as ADE20K \citep{ade20k}, COCO \citep{coco}, and Cityscapes \citep{cityscapes}, which predominantly focus on standard categories like vehicles and buildings. As demonstrated in Section~\ref{sec:experiments}, standard closed-vocabulary segmentation models~\citep{yolov11, mask2former, segformer} omit accessibility-critical classes from their pretrained label spaces entirely, while open-vocabulary and promptable segmentation models~\citep{san, catseg, sed, yoloe, sam3} exhibit severe performance degradation on these categories in zero-shot settings.

Targeted efforts have emerged to address this data scarcity, but their impact remains limited because most operate within narrowly defined, domain-specific scopes that constrain broader downstream deployment. A large body of work has investigated obstacle detection for sidewalk navigation of BLV individuals \citep{sideguide, vidvip, wotr, psid-dataset}, with an emphasis on localization and recognition of static and dynamic obstacles that obstruct safe and continuous walking. Other datasets, such as SurDis \citep{surdis}, target more specialized hazards, particularly surface discontinuities such as steps and drop-offs. GuideTWSI \citep{guidetwsi} focuses narrowly on tactile walking surface indicators. More recent datasets support egocentric scene understanding \citep{sanpo} and accessibility-aware descriptions \citep{guidedogdataset,sait}. These efforts motivate complementary coverage of accessibility-critical objects through dense annotations and controllable synthetic augmentation. \citet{blv-dataset-islam} introduce a taxonomy of 90 BLV-critical object categories with image-level presence labels, rather than the dense pixel-level annotations required for segmentation. As a result, existing real-world datasets remain limited in scale, dense annotations, semantic coverage, and intra-class diversity.

Beyond these structural limitations, a fundamental viewpoint mismatch persists between existing data and deployed assistive perception platforms. Most vision datasets are curated from human-eye-level or vehicle-mounted perspectives, whereas assistive platforms---such as chest-worn wearables \citep{isana-wearable}, cane-mounted cameras \citep{glidance}, and quadruped robots \citep{xiao2021robotic, cai2024navigating, hwang2026guidenav, kim2025understanding}---perceive the world from fundamentally different egocentric viewpoints. This mismatch between training distributions and embodied sensor inputs is further compounded by limited intra-class diversity and insufficient coverage of environmental variations, resulting in poor zero-shot segmentation performance \citep{iod-dataset, blv-survey}.

To bridge these gaps, we develop \textsc{NavAble}, a dataset and a generator tailored to perception for BLV-assistive navigation. \textsc{NavAble} is guided by two principles: first, reliable navigation demands large-scale real-world coverage of fine-grained accessibility-critical object categories selected through prior BLV-related human studies \citep{hwang2026guidenav, islam2024identifying}; second, robust deployment requires a controllable synthetic data generator capable of producing viewpoint-specific data that reflects the geometry, sensing perspective, and environmental variations encountered by assistive platforms in practice. This paper makes two contributions and reports four empirical findings.

\emph{Technical Contributions:}\vspace{-2mm}
\begin{list}{\textbullet}{%
  \setlength{\leftmargin}{8pt}
  \setlength{\itemindent}{8pt}
\setlength{\itemsep}{1pt}
}
\item \textbf{NavAble Dataset.} A large-scale dataset of 41.5K densely annotated real-world images spanning 11 accessibility-critical object classes, comprising 5.5K newly collected egocentric images and 36K curated open-source images aggregated and re-standardized from fragmented class-specific repositories. To the best of our knowledge, this is the largest collection of dense pixel-level annotations targeting BLV-assistive perception.
\item \textbf{NavAble Generator.} A semi-automated synthetic data generation pipeline that transforms uncurated web images into simulation-ready 3D assets and renders them in NVIDIA Isaac Sim with controllable scene composition, viewpoint, and lighting. The pipeline yields 500 class-validated assets across 11 categories and a companion Isaac Sim extension that converts one-time teleoperated trajectories into compositionally scalable supervision, producing 451K densely annotated synthetic frames (NavAble$_{syn}$) with paired RGB, semantic segmentation, depth, surface normals, and 2D/3D bounding boxes.
\end{list}

\emph{Findings:}\vspace{-2mm}
\begin{list}{\textbullet}{%
  \setlength{\leftmargin}{8pt}
  \setlength{\itemindent}{8pt}
\setlength{\itemsep}{1pt}
}
\item \textbf{Benchmark gaps in existing models.} Closed-vocabulary models omit BLV-relevant classes from their fixed pretraining taxonomies, and state-of-the-art open-vocabulary models evaluated in zero-shot settings achieve at most 26.07\% foreground mIoU on accessibility-critical objects, quantifying the data bottleneck that motivates \textsc{NavAble}.
\item \textbf{Strong segmentation capability.} Fine-tuning on the \textsc{NavAble} Dataset across seven segmentation architectures spanning convolutional~\citep{deeplabv3plus, encnet}, transformer~\citep{segformer, segman}, mask-classification~\citep{mask2former, san}, and promptable foundation~\citep{sam3} paradigms reaches up to 71.70\% foreground mIoU, and lifts SAM~3 by +51.43 foreground mIoU over its zero-shot baseline using our collected real data alone.
\item \textbf{Architecture-dependent gains from synthetic augmentation.} Adding \textsc{NavAble}$_{syn}$ to real-only training improves five of the seven evaluated architectures by up to +7.19 foreground mIoU (and up to +9.50 when combined with curated data), whereas DeepLabv3+ and SAM~3 perform best on real-world data alone.
\item \textbf{Relative transfer to embodied viewpoints.} Synthetic augmentation yields positive relative transfer on out-of-distribution egocentric inputs from two BLV-assistive platforms (a cane-shaped robot~\cite{glidance} and a quadruped robot), while underscoring that absolute accuracy at 16--23\% mIoU is not yet sufficient for autonomous deployment.
\end{list}

\begin{wrapfigure}{r}{0.45\linewidth}
\vspace{-1.1em}
\centering
\includegraphics[width=0.9\linewidth]{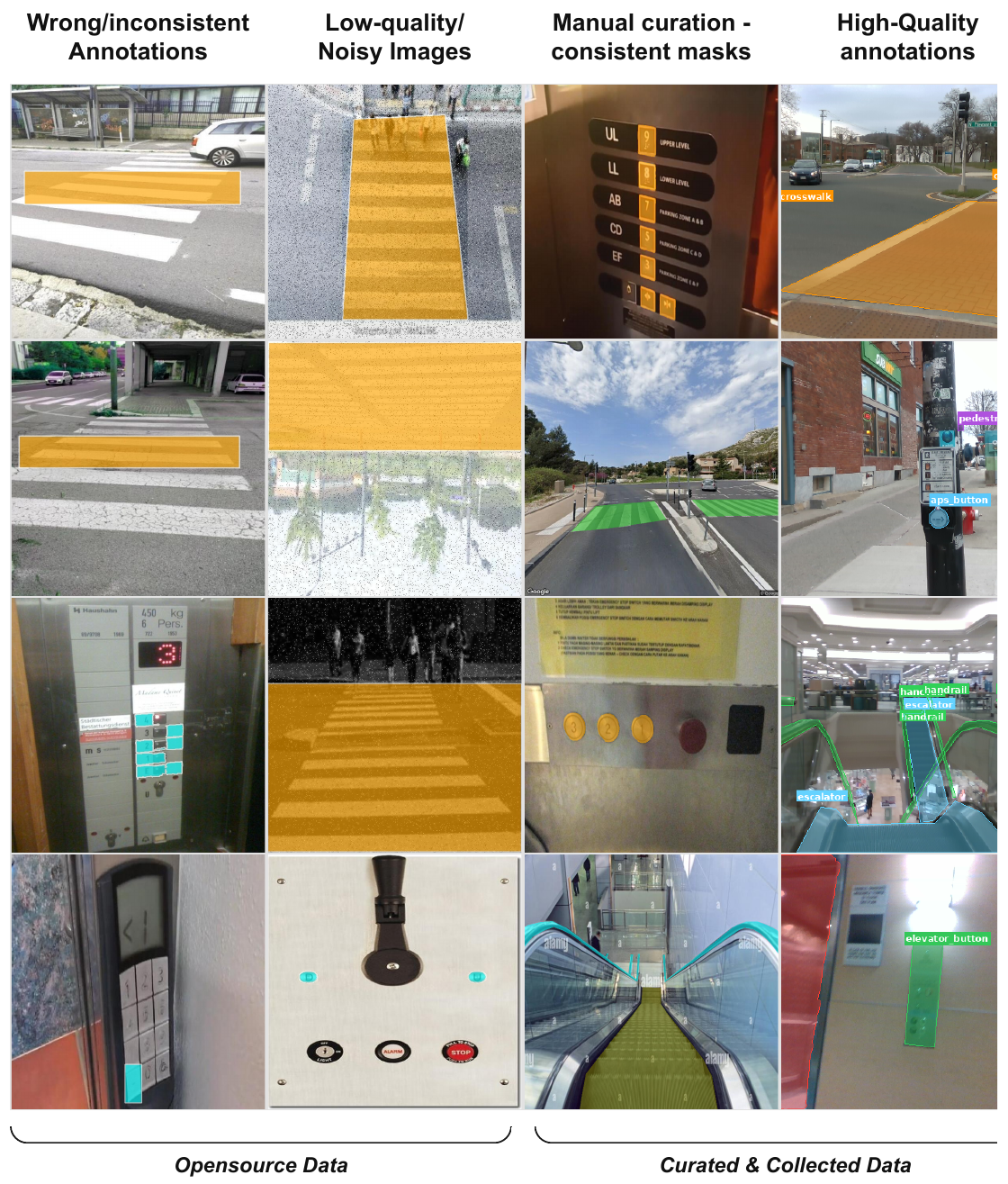}
\caption{Inconsistent open-source annotations (left) are curated and merged with our densely annotated real-world data (right).}
\label{fig:data-curation}
\vspace{-1.0em}
\end{wrapfigure}

\section{Real-world data: curation and collection}
In \textsc{NavAble}, we focus on 11 accessibility-critical object classes chosen based on three criteria: (1) high navigational utility, (2) frequent occurrence in assistive settings, and (3) limited coverage in existing datasets. The selected 11 classes correspond to objects that directly inform mobility decisions --- such as crossing streets, entering buildings, locating transit infrastructure, or navigating vertical transitions --- while spanning both indoor and outdoor environments.

We first aggregate annotated images from multiple open-source data hosting platforms (e.g., Roboflow, Hugging Face), the majority of which contain single-class coverage of accessibility-related objects. While these datasets provide useful starting points, we identify several systematic limitations that constrain their direct use: 1) \textbf{Inconsistent and imprecise annotations:} Annotation standards vary across sources in both granularity and definition, and segmentation masks often exhibit loose or inaccurate boundaries, as shown in Figure~\ref{fig:data-curation},  2) \textbf{Severe class imbalance:} The distribution across categories is highly skewed, with a few classes heavily over-represented in open-source data (e.g., 27.8K crosswalk, 6.7K pedestrian signal, and 4.4K elevator button images, corresponding to 49.4K, 19.4K, and 39.0K instances) while others are entirely absent (e.g., revolving door, door button, bus stop, bus stop sign, APS button), as shown in Figure~\ref{fig:overview}, 3) \textbf{Fragmented semantic coverage:} The majority of datasets are single-class or narrowly scoped (e.g., only bus stop without signs), lacking the co-existing context necessary for robust scene understanding, and 4) \textbf{Class ambiguity and noise:} Mislabeled or visually ambiguous samples introduce noise and degrade training stability. These issues motivated us to perform both substantial curation and targeted real-world collection. 

\subsection{Data curation from open-source datasets}
\label{curated-data}

To mitigate the issues of annotation noise, class inconsistency, and unreliable supervision, we perform a rigorous curation process that yields approximately 36K annotated images. First, we standardize class definitions across datasets to align with our target taxonomy of 11 accessibility-critical classes. We then carry out careful verification to filter out low-quality and label-inconsistent samples. While the severe class imbalance problem was unavoidable, this process produces a curated dataset with precise annotations and consistent multi-class semantics, forming reliable real-world data for training. Curated repositories, the filtering criteria and exclusion statistics are detailed in Appendix~\ref{app:annotation_filtering} and ~\ref{app:curation_sources}.

\subsection{Targeted real-world data collection}
\label{collected-data}
Despite extensive curation, several accessibility-critical categories remain underrepresented or entirely absent in existing datasets. Moreover, inconsistencies in mask definitions cannot be fully resolved without reannotation. To address the limitations of currently available open-source data, we undertake targeted real-world data collection and manually annotate dense pixel-level labels across our target taxonomy (covering 10 classes in the primary chest-mounted split and 100 supplementary \texttt{revolving\_door} frames; Appendix~\ref{app:revolving_door}). The collection process is designed to capture (i) \textbf{diverse environments}, spanning indoor and outdoor settings relevant to navigation; (ii) \textbf{embodiment-relevant viewpoints}, i.e., egocentric perspectives aligned with assistive systems; and (iii) \textbf{contextual co-occurrence}, with objects embedded within realistic scene contexts.

We collected data using a custom capture rig equipped with an Intel RealSense D435 mounted at chest height to approximate an egocentric assistive viewpoint. Data are recorded across 50 different locations, spanning a wide range of viewpoints and both indoor (e.g., campus building) and outdoor environments (e.g., suburban area). Collection locations are visualized in Fig.~\ref{fig:coverage}
(Appendix~\ref{app:annotation_filtering}). To ensure high data quality, we first manually filtered noisy/blurry samples. We considered maximizing diversity across viewpoints, environments, lighting conditions, and intra-class variations. We then annotate the remaining data using Label Studio~\cite{Labelstudio}, producing polygon-level semantic segmentation labels in COCO-style format for all target object classes. We identify annotation errors and missing instances through a two-stage verification process, followed by iterative correction and refinement. The final annotations are obtained after three full passes over the collected data. Each image is densely annotated in a multi-class setting, labeling all visible instances of the target categories within the frame, as depicted in the right column of Figure~\ref{fig:data-curation}. Dataset statistics, including per-class counts, are summarized in Figure~\ref{fig:overview}. Appendix~\ref {app:annotation_filtering} details the annotation protocol, multi-pass verification, disagreement handling, and independent agreement audit.

\section{Scalable synthetic data generation (SDG) pipeline} 
\label{sdg-pipeline}

Real-world data alone cannot provide the scale, diversity, and viewpoint variability required for robust perception of accessibility-critical objects, due to sparse coverage of rare classes, limited viewpoints, and high annotation cost. To address this fundamental bottleneck, we introduce a scalable synthetic data generation pipeline built on NVIDIA Isaac~Sim that produces large-scale, densely annotated data with controllable scene composition and viewpoint diversity. While the \textsc{NavAble} dataset integrates curated and newly collected real-world data, rare accessibility-critical categories remain inherently difficult to capture at scale, making synthetic data a necessary complement for coverage and generalization. We adopt Isaac~Sim for its dynamics-aware simulation capabilities and potential to support learning across diverse robotic embodiments; in this work, we focus on its impact on training perception models.

\subsection{From web images to simulation-ready 3D assets using foundation models}
\label{sec:navable_pipeline}
\begin{figure}[t]
\centering
\includegraphics[width=\linewidth]{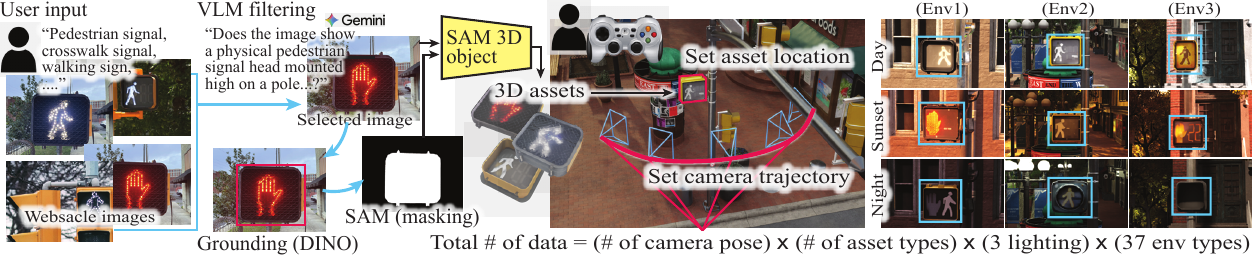}
\caption{\textbf{Overview of \textsc{NavAble} Generator}. (Left) Based on the user's object of interest, we generate 3D assets from images by using foundation models. (Center) Our generator only requires two pieces of information from the user: the asset location and the camera trajectory. Once those are given, the generator creates augmented data. (Right) \textsc{NavAble} generator creates data in diverse scenes, lighting conditions, and environments via Isaac Sim.}
\vspace{-3mm}
\label{fig:overview-sdg}
\end{figure}

We construct a large and diverse asset library through our custom-developed pipeline that transforms web-scale image data into simulation-ready 3D meshes. 
Given class-specific keyword queries, images are collected from multiple search engines (e.g., Wikimedia Commons). Because keyword-based retrieval introduces substantial noise and weak semantic alignment, we use a vision-language model gemini-3-flash-preview to filter out unsuitable samples, including diagrams, CAD renderings, and heavily occluded instances. For each filtered image, we localize objects of interest using Grounding DINO~\cite{liu2023grounding} and suppress duplicate detections via non-maximum suppression. The remaining bounding boxes are then passed to SAM2.1~\cite{ravi2024sam2} to produce precise segmentation masks. We aggregate per-box masks into a single coherent object mask, which is critical for reconstructing composite objects (e.g., APS buttons with both signage and tactile interfaces) as structurally consistent assets. The resulting image–mask pairs are reconstructed into textured 3D meshes via SAM 3D Objects~\cite{sam3dteam2025sam3d3dfyimages} and exported in Universal Scene Description (USD) format for direct integration into Isaac~Sim as shown in Fig.~\ref{fig:overview-sdg}.

This pipeline yields over 500 class-validated assets across 11 categories. As initial image and mask screening is fully automated, human effort during asset curation is restricted to final 3D mesh inspection. In a measured 150-image pilot audit across three classes (APS button, escalator, and pedestrian signal), automated VLM validation removed 105 of 148 crawled images and automated mask-quality filtering removed a further 9 of 43 candidates without human involvement; researchers inspected only the 34 surviving 3D meshes at a measured rate of approximately 1~minute per 10~meshes, retaining 24 usable assets (16.0\% overall yield; see Appendix~\ref{app:synth_details} for stage-wise yield and manual-effort statistics). After manual verification, we retain the highest-quality assets per class for simulation, while releasing the full asset set to support future research. Importantly, SAM~3D is used once offline solely to reconstruct 3D asset meshes rather than to generate segmentation labels; while foundation-model mask errors can degrade asset geometry, all synthetic ground-truth annotations are read directly from the Isaac~Sim scene graph at render time and are therefore pixel-exact by construction.

Once assets are generated, our \textsc{NavAble} Isaac~Sim extension enables users to record camera trajectories via simple teleoperation (e.g., using a remote controller). Each trajectory is stored as a sequence of poses and, given a fixed object placement (6D pose and scale) within a scene, can be replayed across different environments, lighting conditions, and asset styles. This design enables compositional scaling of synthetic data, effectively converting interactive data collection into a one-time cost. All annotations are derived directly from the scene graph, ensuring perfect consistency under variations in assets, lighting, and environments. In practice, 10--20~minutes of teleoperated trajectory recording yields densely annotated segmentation data that would otherwise require 5--6~hours of manual collection and pixel-level labeling. For instance, 5 user-recorded trajectories replayed across 4 environments and 10 asset variants produce 20{,}000 annotated frames (and up to 60{,}000 frames when iterated across 3 lighting variations), while post-generation screening to remove frames without visible target annotations ($<$1\% of rendered output) requires approximately 1~minute per 100~images. We release this extension together with a companion command-line interface, enabling practitioners to efficiently generate large-scale, densely annotated datasets across new object classes, environments, and viewpoints with minimal additional effort.

\subsection{NavAble synthetic dataset}
\label{sec:synth-dataset}

\newcommand{\wratio}{0.14}
\begin{table}[t]
\centering
\footnotesize
\setlength{\tabcolsep}{0.5pt}
\renewcommand{\arraystretch}{1.15}
\setlength{\aboverulesep}{0pt}
\setlength{\belowrulesep}{1pt}
\caption{\textbf{Quantitative metrics and qualitative comparison across segmentation models.}
10-class foreground-only mIoU (\textbf{FG mIoU}), 11-class background-inclusive mIoU (\textbf{mIoU w/ bg}), foreground precision, and foreground recall on the held-out \textsc{NavAble}-Real test split ($N{=}1{,}482$) for the representative checkpoints visualized below (full configuration sweep across all seven architectures in Table~\ref{tab:main-results-comparison}). The lower block shows ground-truth masks (left) and each model's prediction in the same column order.}
\begin{adjustbox}{max width=\textwidth}
\begin{tabular}{c |c c c c c c}
\toprule
\textbf{Metric} & \textbf{SAN-ZS} & \textbf{SAN} & \textbf{DeepLabv3+}
                & \textbf{EncNet} & \textbf{SegFormer} & \textbf{Mask2Former} \\
\midrule
FG mIoU      & 14.94 & 66.96 & 60.88 & 57.24 & 63.15 & \textbf{70.33} \\
mIoU (w/ bg) & 22.10 & 69.45 & 63.84 & 60.56 & 65.96 & \textbf{72.67} \\
Precision    & 31.17 & 82.53 & 69.36 & 64.68 & 70.18 & \textbf{83.73} \\
Recall       & 47.10 & 75.45 & 82.43 & 81.16 & \textbf{84.37} & 79.24 \\
\toprule
\includegraphics[width=\wratio\textwidth, keepaspectratio]{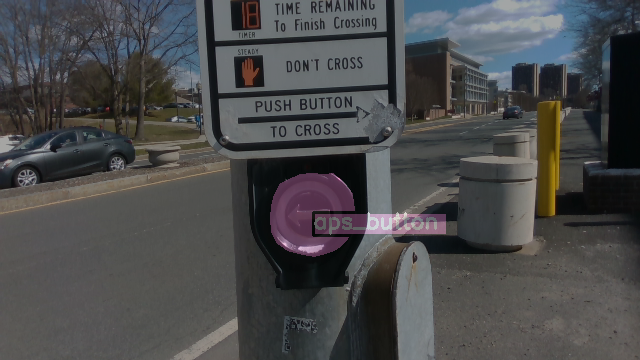}
  & \includegraphics[width=\wratio\textwidth, keepaspectratio]{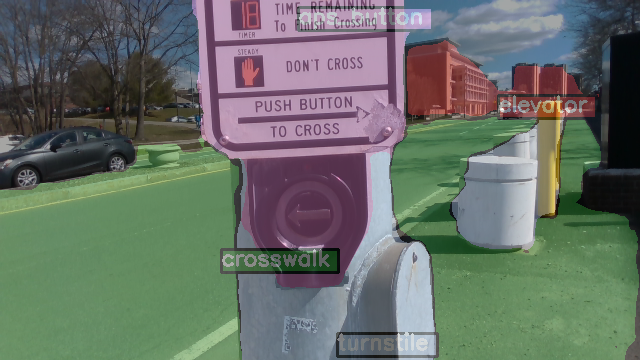}
  & \includegraphics[width=\wratio\textwidth, keepaspectratio]{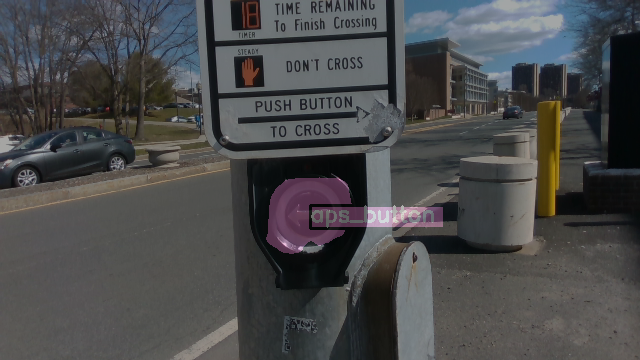}
  & \includegraphics[width=\wratio\textwidth, keepaspectratio]{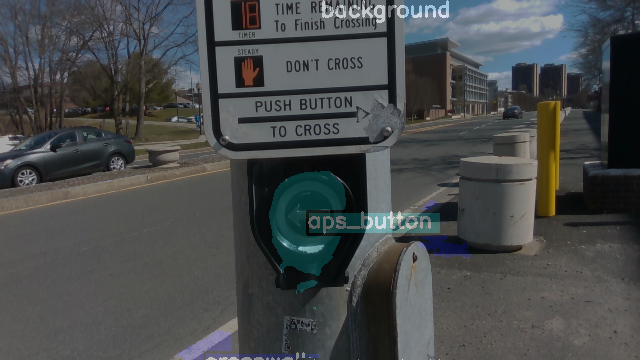}
  & \includegraphics[width=\wratio\textwidth, keepaspectratio]{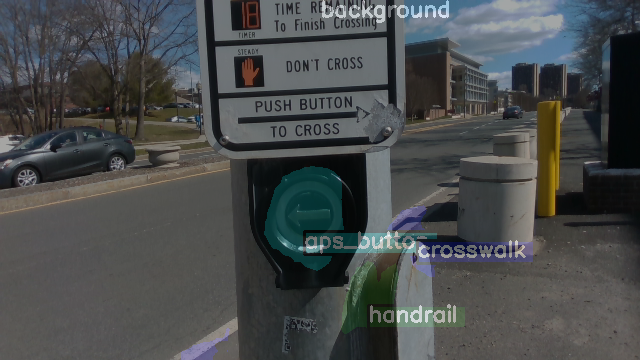}
  & \includegraphics[width=\wratio\textwidth, keepaspectratio]{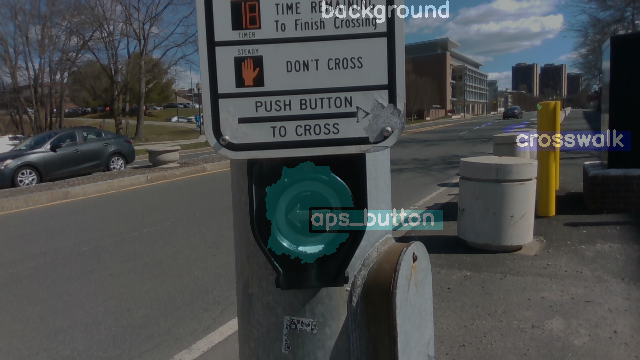}
  & \includegraphics[width=\wratio\textwidth, keepaspectratio]{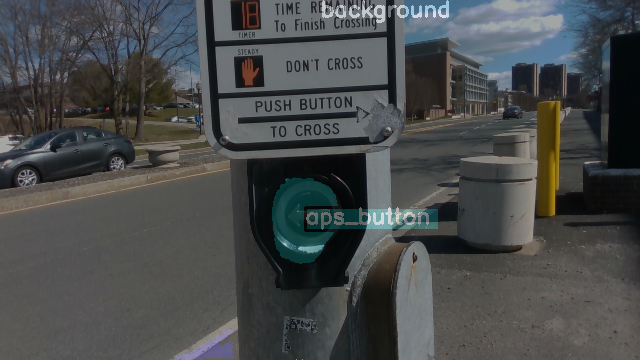} \\
\includegraphics[width=\wratio\textwidth, keepaspectratio]{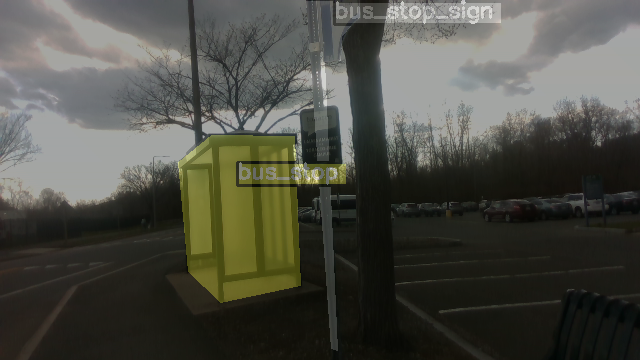}
  & \includegraphics[width=\wratio\textwidth, keepaspectratio]{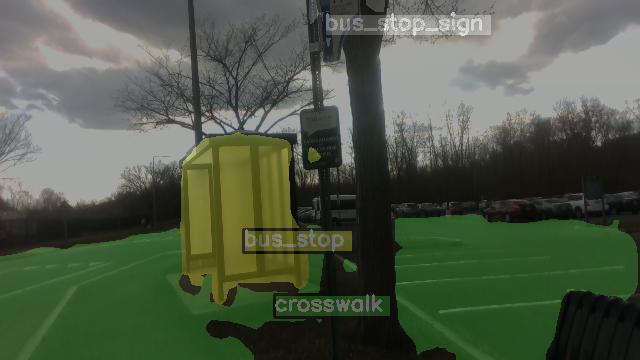}
  & \includegraphics[width=\wratio\textwidth, keepaspectratio]{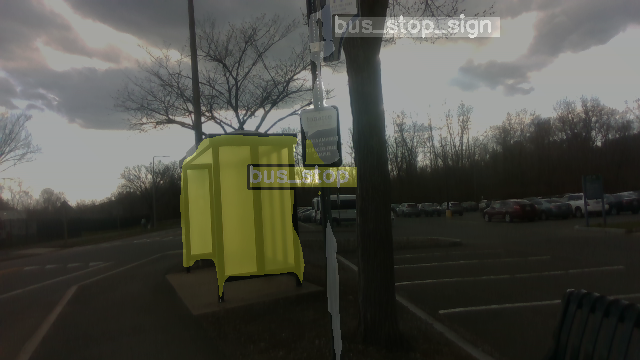}
  & \includegraphics[width=\wratio\textwidth, keepaspectratio]{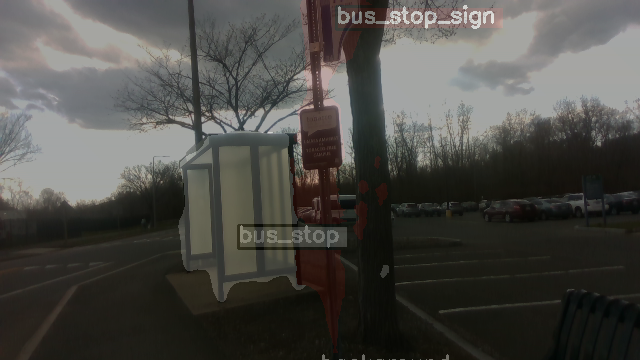}
  & \includegraphics[width=\wratio\textwidth, keepaspectratio]{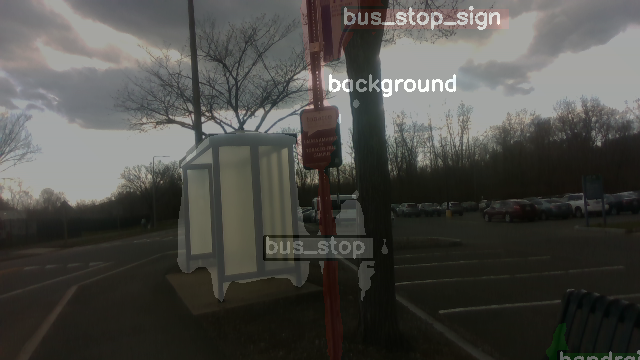}
  & \includegraphics[width=\wratio\textwidth, keepaspectratio]{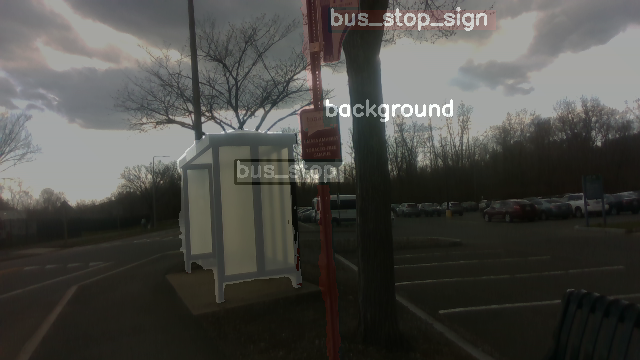}
  & \includegraphics[width=\wratio\textwidth, keepaspectratio]{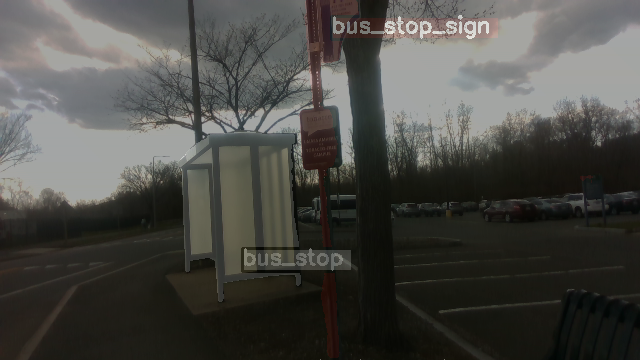} \\
\includegraphics[width=\wratio\textwidth, keepaspectratio]{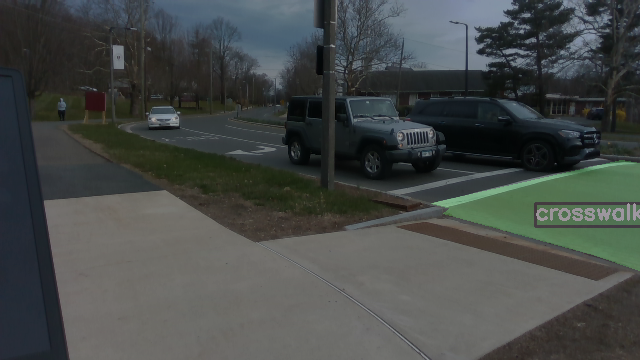}
  & \includegraphics[width=\wratio\textwidth, keepaspectratio]{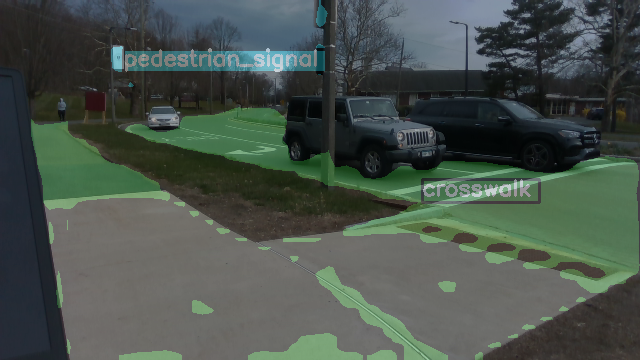}
  & \includegraphics[width=\wratio\textwidth, keepaspectratio]{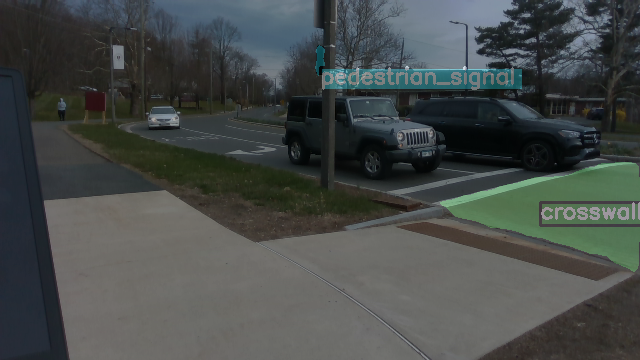}
  & \includegraphics[width=\wratio\textwidth, keepaspectratio]{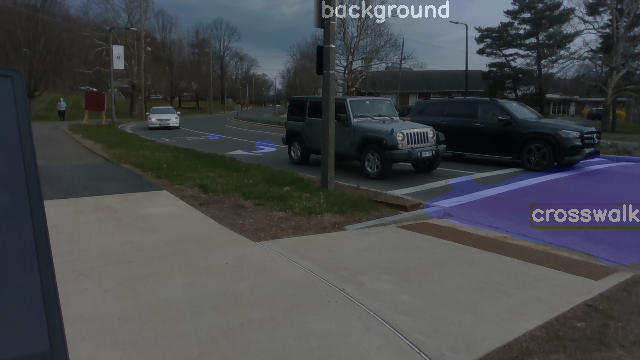}
  & \includegraphics[width=\wratio\textwidth, keepaspectratio]{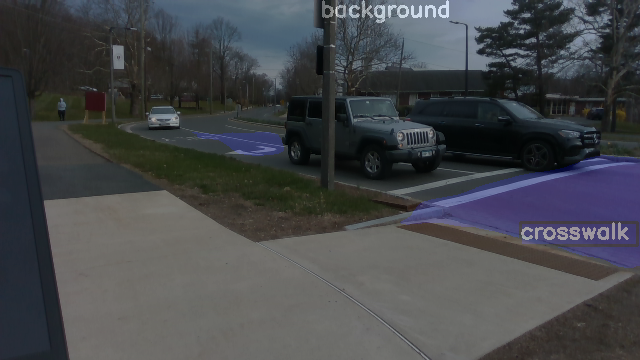}
  & \includegraphics[width=\wratio\textwidth, keepaspectratio]{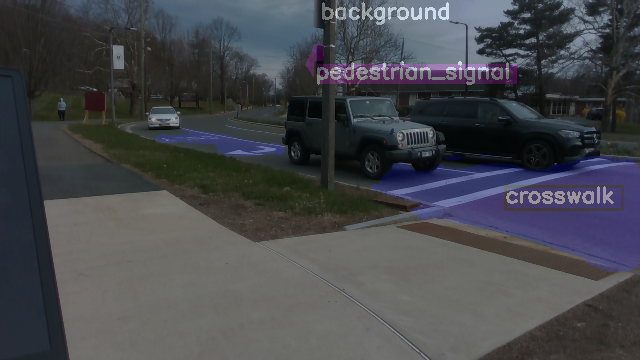}
  & \includegraphics[width=\wratio\textwidth, keepaspectratio]{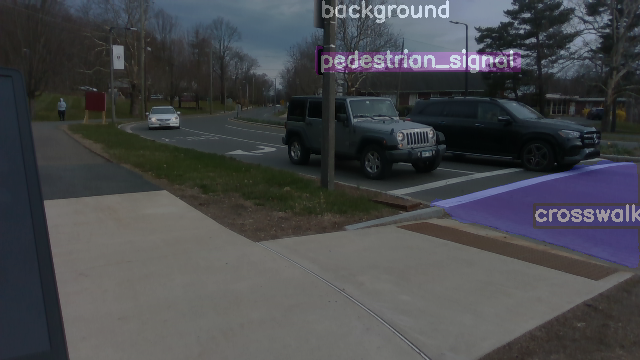} \\
\includegraphics[width=\wratio\textwidth, keepaspectratio]{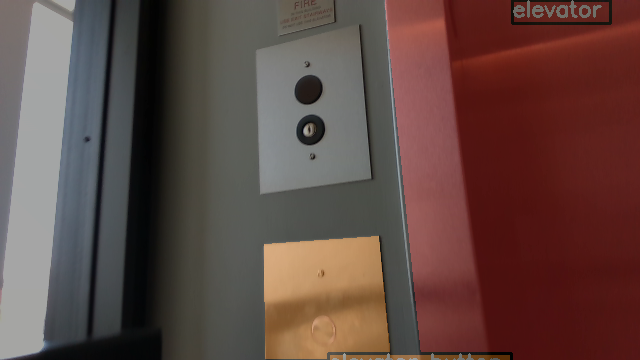}
  & \includegraphics[width=\wratio\textwidth, keepaspectratio]{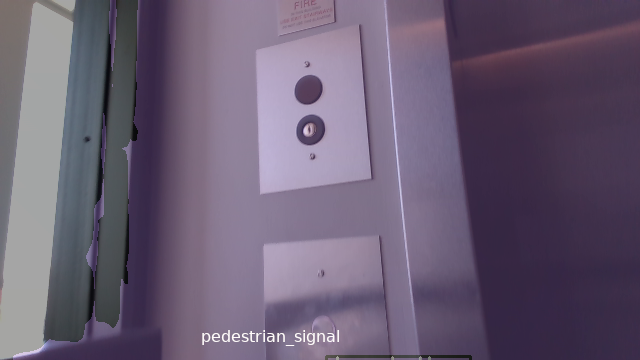}
  & \includegraphics[width=\wratio\textwidth, keepaspectratio]{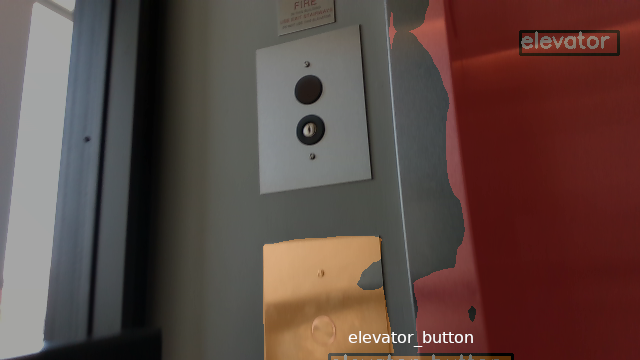}
  & \includegraphics[width=\wratio\textwidth, keepaspectratio]{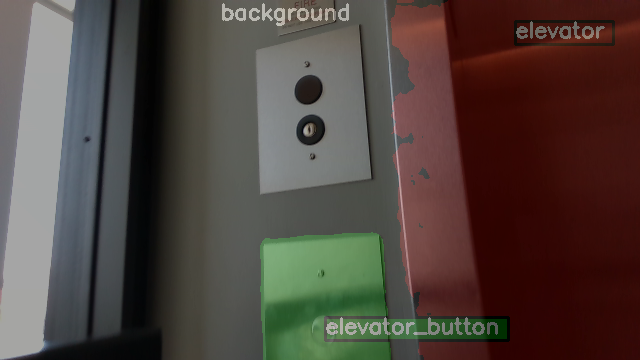}
  & \includegraphics[width=\wratio\textwidth, keepaspectratio]{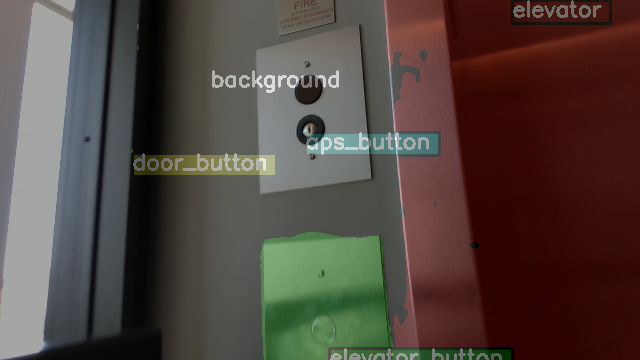}
  & \includegraphics[width=\wratio\textwidth, keepaspectratio]{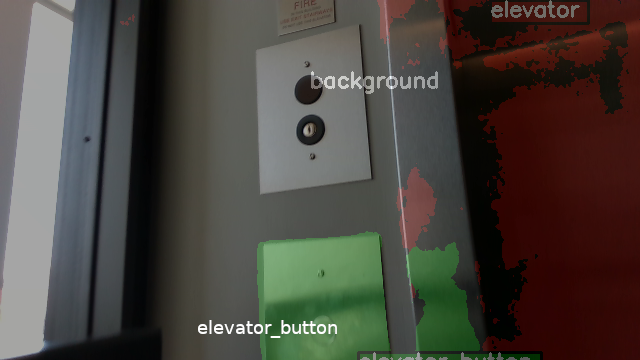}
  & \includegraphics[width=\wratio\textwidth, keepaspectratio]{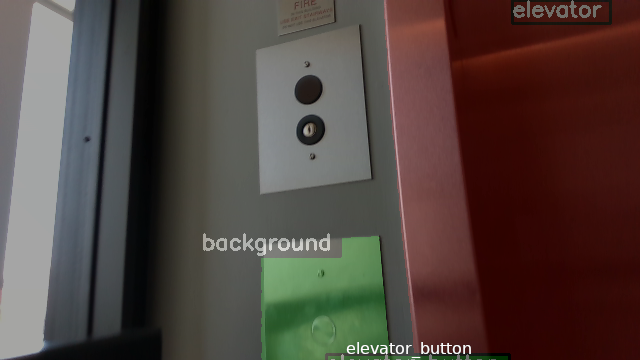} \\
\footnotesize Ground Truth
  & \footnotesize SAN-ZS
  & \footnotesize SAN
  & \footnotesize DeepLabv3+
  & \footnotesize EncNet
  & \footnotesize SegFormer
  & \footnotesize Mask2Former \\
\bottomrule
\end{tabular}
\end{adjustbox}
\label{tab:exp-highlight}
\vspace{-4mm}
\end{table}

Using our extension, we construct the synthetic split of \textsc{NavAble}, denoted as \textsc{NavAble}$_{syn}$. We curate 37 3D environments from the Fab marketplace~\citep{fab2026}, selected for visual realism and contextually appropriate placement of target classes (e.g., elevator buttons in lobbies, bus-stop signs along sidewalks). Within these environments, we define 488 locations, each populated with one of the top-10 retained assets per class from Sec.~\ref{sec:navable_pipeline}. We sampled every 50 frames during offline rendering. The resulting dataset contains over \textbf{25K} trajectories, \textbf{451K} RGB frames (450{,}860 rendered frames), and \textbf{781K} instance annotations across the eleven \textsc{NavAble} classes, as shown in Figure~\ref{fig:overview}. Each frame is a pixel-aligned multi-modal tuple comprising RGB, semantic segmentation, 2D and 3D bounding boxes, surface normals, and depth. For experiments, we use a subset of the data (detailed in Sec.~\ref{subsec:setup}), while releasing the full dataset to the community to support future research.
\vspace{-1mm}
\section{Experiments}
\label{sec:experiments}
\vspace{-2mm}
We evaluate \textsc{NavAble} along four axes. First, we assess whether existing state-of-the-art segmentation systems can recognize BLV-relevant object classes in zero-shot settings. Second, we test the extent to which fine-tuning on our collected real-world data improves segmentation performance. Third, we study whether augmenting training with large-scale synthetic data from our SDG pipeline yields additional gains beyond real data alone. Finally, we evaluate whether these improvements transfer to out-of-distribution egocentric viewpoints on two mobility-assistive platforms.
\vspace{-2mm}

\subsection{Experimental setup}
\vspace{-2mm}
\label{exp-setup}
The \textsc{NavAble} dataset, covering the 11 target BLV classes, is drawn from three complementary sources: curated open-source data (\textsc{NavAble}\(_{\text{cur}}\)), collected real-world data (\textsc{NavAble}\(_{\text{col}}\)), and our SDG-pipeline-generated synthetic data (\textsc{NavAble}\(_{\text{syn}}\)). The dataset is partitioned into training, validation, and test splits, where the held-out test split (1{,}482 images) is drawn exclusively from real-world data (\textsc{NavAble}\(_{\text{col}}\)) to ensure unbiased evaluation of real-world generalization. In addition, we enforce location-level separation across splits, such that all test images come from locations excluded from the training and validation sets. Model performance is reported in terms of mean Intersection-over-Union (mIoU), precision, and recall. To prevent metric inflation from the high-accuracy \texttt{background} class ($\sim$90--96\% IoU), our main benchmark tables report 10-class foreground-only metrics over the evaluated target categories, while 11-class background-inclusive mIoU (\textbf{mIoU w/ bg}) is reported alongside foreground mIoU in Table~\ref{tab:exp-highlight} and in the per-class tables of Appendix~\ref{app:per_class}. We omit mAP as a universal metric since pixel-classification models (SegFormer, SegMAN, DeepLabv3+, EncNet) produce no instance-level confidence scores required for its computation.

\begin{table}[!htbp]
\centering
\small
\setlength{\tabcolsep}{6pt}
\caption{\textbf{Open-vocabulary zero-shot baselines on the held-out \textsc{NavAble}-Real test split} ($N{=}1{,}482$ images; \texttt{revolving\_door} excluded). Each model is evaluated at its best configuration from a hyperparameter sweep; \texttt{+bg} indicates an explicit background prompt vocabulary. \textbf{Best} per column in bold.}
\label{tab:zs-baselines}
\begin{tabular}{l l cccc}
\toprule
\textbf{Model} & \textbf{Backbone / Setting} & \textbf{mIoU} & \textbf{mAP$_{50:95}$} & \textbf{Prec} & \textbf{Rec} \\
\midrule
SAN-ZS~\citep{san}      & CLIP ViT-B/16         & 14.94          & \phantom{0}3.84 & 31.17          & 47.10 \\
YOLOE~\citep{yoloe}     & YOLO11-M (conf 0.05)  & 14.94          & \phantom{0}2.32 & 29.07          & 40.33 \\
SAM~3~\citep{sam3}      & SAM~3 (conf 0.25)     & 19.06          & \phantom{0}1.80 & \textbf{51.59} & 34.17 \\
SED~\citep{sed}         & ConvNeXt-L (+bg)      & 24.07          & \phantom{0}9.50 & 38.39          & \textbf{47.90} \\
CAT-Seg~\citep{catseg}  & ViT-L/14 (+bg)        & \textbf{26.07} & \textbf{12.21}  & 40.72          & 46.83 \\
\bottomrule
\end{tabular}
\vspace{-2mm}
\end{table}

\subsubsection{Model training and setup}
\label{subsec:setup}

\textbf{Architectures.}
We benchmark seven open-source segmentation models spanning modern semantic segmentation paradigms: CNN per-pixel baselines DeepLabv3+~\citep{deeplabv3plus} and EncNet~\citep{encnet} (both ResNet-101, output stride $8$); hierarchical transformer and state-space-augmented per-pixel models SegFormer~\citep{segformer} (MiT-B5) and SegMAN~\citep{segman}; query-based mask-classification model Mask2Former~\citep{mask2former} (Swin-L); open-vocabulary mask-classification model SAN~\citep{san} (CLIP ViT-B/16); and promptable foundation model SAM~3~\citep{sam3}. Models are initialized from publicly released ADE20K weights (COCO-Stuff-164K for SAN; official pretrained checkpoint for SAM~3) and fine-tuned end-to-end for our 12-class label space ($1$ background $+\,11$ accessibility-critical foreground classes).

\textbf{Training-data configurations.}
For each architecture we evaluate \emph{four} progressively richer configurations: a real-only baseline (\emph{Real}), a real$+$curation configuration (\emph{Real+Cur}), a real$+$synthetic ablation that isolates synthetic augmentation without curation (\emph{Real+Syn}), and a full combined configuration (\emph{Real+Cur+Syn}). Synthetic budgets are matched by foreground \emph{instances} per class rather than image count, since several classes (e.g., elevators and elevator buttons) frequently co-occur in the same frame. The validation and test splits are held identical across all four configurations. 

Because \texttt{revolving\_door} has no positive real instances in the primary \emph{Real} or \emph{Cur} splits (and is supervised only in \emph{Syn}), the 1{,}482-image benchmark evaluates the 10 real-annotated foreground classes (see Appendix~\ref{app:per_class}), while \texttt{revolving\_door} is evaluated on a separate 100-frame set in Appendix~\ref{app:revolving_door}. Training schedules are detailed in Appendix~\ref{app:training_details}.

\subsection{Evaluation results}
\label{sec:eval-results}
\textbf{Can state-of-the-art models segment target BLV classes in zero-shot settings?}

\begin{wraptable}{r}{0.62\linewidth}
\vspace{-1.0em}
\centering
\footnotesize
\setlength{\tabcolsep}{3.5pt}
\renewcommand{\arraystretch}{1.15}
\caption{\textbf{VLM evaluation on \textsc{NavAble}.}
363 real-world test images, 10 categories (excluding revolving door).
All metrics are percentages. Box AP follows COCO conventions;
Pt-in-mask measures predicted points inside ground-truth masks;
PCK thresholds are fractions of the image diagonal.
MolmoPoint outputs points only. Best per column in \textbf{bold}.}
\label{tab:vlm-main}
\begin{adjustbox}{max width=\linewidth}
\begin{tabular}{l cc ccc}
\toprule
& \multicolumn{2}{c}{\textbf{Box}}
& \multicolumn{3}{c}{\textbf{Point}} \\
\cmidrule(lr){2-3}\cmidrule(lr){4-6}
\textbf{Model} & AP & AP$_{50}$ & Pt-in-mask & PCK$_{.05}$ & PCK$_{.10}$ \\
\midrule
Gemini Robotics-ER 1.6
& \textbf{33.8} & \textbf{48.9} & 69.0 & \textbf{68.7} & \textbf{85.5} \\
Gemini 3 Pro
& 31.5 & 45.9 & 68.6 & 61.2 & 79.4 \\
Qwen3-VL-32B
& 20.0 & 29.7 & 59.0 & 59.7 & 79.2 \\
MolmoPoint-8B
& -- & -- & \textbf{74.2} & 55.0 & 73.6 \\
\bottomrule
\end{tabular}
\end{adjustbox}
\vspace{-0.5em}
\end{wraptable}

We first evaluate out-of-the-box segmentation capability on the target BLV classes. We explicitly distinguish between closed-vocabulary and open-vocabulary evaluation protocols: closed-vocabulary architectures considered in this work (Mask2Former, SegFormer, DeepLabv3+, and EncNet) are pretrained on fixed label spaces (e.g., ADE20K, COCO) that do not contain most accessibility-critical categories. Evaluating them zero-shot on unseen classes would be uninformative, so we evaluate closed-vocabulary models exclusively after fine-tuning on \textsc{NavAble} (Table~\ref{tab:exp-highlight} and Table~\ref{tab:main-results-comparison}). To establish a fair zero-shot baseline on the full 1{,}482-image test split, we evaluate five open-vocabulary and promptable segmentation models spanning recent architectures: SAN~\citep{san}, CAT-Seg~\citep{catseg}, SED~\citep{sed}, YOLOE~\citep{yoloe}, and SAM~3~\citep{sam3} (Table~\ref{tab:zs-baselines}). Each zero-shot baseline is reported at its best hyperparameter configuration from a validation sweep, with CAT-Seg and SED additionally provided an explicit background vocabulary that improves their scores by 10--15 mIoU. Even under these favorable settings, the strongest zero-shot model (CAT-Seg at 26.07\% mIoU) falls far below every fine-tuned configuration on \textsc{NavAble} (48.88\%--71.70\% foreground mIoU in Table~\ref{tab:main-results-comparison}). The zero-shot shortfall is particularly severe on small or thin accessibility-critical long-tail classes: across all five zero-shot models, \texttt{aps\_button} peaks at 2.92\% IoU (2.92\% for SAN-ZS, and 0.36\%--1.46\% across the four newer models in Appendix~\ref{app:per_class}, Table~\ref{tab:perclass-zs}), \texttt{handrail} at 15.05\%, and \texttt{door\_button} at 16.34\%. Moreover, the 2025 architectures (SAM~3 at 19.06\% and YOLOE at 14.94\% mIoU) do not outperform earlier open-vocabulary methods, confirming that the BLV perception gap is not resolved by model recency or pretraining scale alone, but requires task-specific supervision.

For reference, we evaluate state-of-the-art proprietary and open-weights vision-language models (Gemini Robotics-ER 1.6, Gemini 3 Pro, Qwen3-VL-32B, MolmoPoint-8B) in Table~\ref{tab:vlm-main}. However, we treat these as indicative reference points rather than deployable baselines: cloud-API reliance is incompatible with the always-on, low-latency, privacy-preserving perception BLV navigation requires, and closed weights or high compute requirements preclude on-device deployment.

\textbf{Does fine-tuning on \textsc{NavAble} real-world data establish a strong baseline?}
To overcome the limitations of zero-shot inference, we fine-tune all seven segmentation architectures on our collected (\emph{Real}) and curated (\emph{Real+Cur}) real-world data (Table~\ref{tab:main-results-comparison}). Fine-tuning on \textsc{NavAble}\(_{\text{col}}\) (\emph{Real}) alone establishes a strong baseline across every model family, reaching 57.24\%--70.49\% foreground mIoU---more than doubling the strongest zero-shot baseline (CAT-Seg at 26.07\%). On the exact same architecture, fine-tuning SAM~3 on \emph{Real} lifts foreground mIoU from 19.06\% (zero-shot) to \textbf{70.49\%} (\textbf{+51.43 mIoU}). Augmenting \emph{Real} with curated open-source data (\emph{Real+Cur}) further improves multi-scale transformer and mask-classification models---lifting SegMAN to 69.78\% (+7.58 mIoU), Mask2Former to 68.85\% (+2.70), SegFormer to 59.01\% (+1.20), and SAN to 62.52\% (+0.45)---whereas DeepLabv3+, EncNet, and SAM~3 decline, which we hypothesize may reflect sensitivity to severe class skew and boundary heterogeneity across public sources.

\begin{table*}[t]
\centering
\footnotesize
\setlength{\tabcolsep}{3.5pt}
\renewcommand{\arraystretch}{1.18}
\caption{\textbf{Quantitative results across training-data configurations on the held-out \textsc{NavAble}-Real test split ($N{=}1{,}482$).}
All columns report 10-class foreground-only precision (\textbf{Prec}), recall (\textbf{Rec}), and mIoU (excluding \texttt{background} and \texttt{revolving\_door}). The two rightmost columns report the foreground mIoU change for \textbf{Syn vs.\ Real} (\emph{Real+Syn} minus \emph{Real}) and \textbf{Best Aug.\ vs.\ Real} (best of \emph{Real+Cur}, \emph{Real+Syn}, \emph{Real+Cur+Syn} minus \emph{Real}). Background-inclusive mIoU (up to 73.77\% on SegMAN and 72.67\% on Mask2Former) and per-class breakdowns are reported in Appendix~\ref{app:per_class}. Best per column in \textbf{bold}.}
\begin{adjustbox}{max width=\textwidth}
\begin{tabular}{l rrr rrr rrr rrr cc}
\toprule
& \multicolumn{3}{c}{\textbf{Real}}
& \multicolumn{3}{c}{\textbf{Real + Cur}}
& \multicolumn{3}{c}{\textbf{Real + Syn}}
& \multicolumn{3}{c}{\textbf{Real + Cur + Syn}}
& \multicolumn{2}{c}{\textbf{$\Delta$FG mIoU}} \\
\cmidrule(lr){2-4} \cmidrule(lr){5-7} \cmidrule(lr){8-10} \cmidrule(lr){11-13} \cmidrule(lr){14-15}
\textbf{Model}
& Prec & Rec & mIoU
& Prec & Rec & mIoU
& Prec & Rec & mIoU
& Prec & Rec & mIoU
& Syn vs.\ R & Best Aug.\ vs.\ R \\
\midrule
SegMAN~\citep{segman}           & \textbf{87.94} & 68.64          & 62.20          & \textbf{86.88} & 78.15          & \textbf{69.78} & \textbf{83.74} & 78.45          & 69.39          & \textbf{85.37} & 80.34          & \textbf{71.70} & \textbf{+7.19} & \textbf{+9.50} \\
Mask2Former~\citep{mask2former} & 80.33          & 76.24          & 66.15          & 80.00          & 81.42          & 68.85          & 83.73          & 79.24          & \textbf{70.33} & 79.07          & 81.61          & 68.26          & +4.18          & +4.18 \\
SAM~3~\citep{sam3}              & 84.97          & 80.91          & \textbf{70.49} & 83.88          & 78.19          & 67.42          & 76.52          & 83.52          & 67.48          & 76.60          & 80.87          & 65.35          & $-$3.01        & $-$3.01 \\
SAN~\citep{san}                 & 82.67          & 68.35          & 62.07          & 77.10          & 74.17          & 62.52          & 82.53          & 75.45          & 66.96          & 82.75          & 76.93          & 68.12$^\dagger$& +4.89          & +6.05$^\dagger$ \\
SegFormer~\citep{segformer}     & 64.37          & 81.77          & 57.81          & 64.63          & \textbf{83.84} & 59.01          & 70.42          & 83.00          & 62.76          & 70.18          & 84.37          & 63.15          & +4.95          & +5.34 \\
EncNet~\citep{encnet}           & 64.68          & 81.16          & 57.24          & 62.92          & 66.63          & 48.88          & 66.57          & 83.07          & 59.58          & 62.95          & 85.74          & 58.43          & +2.34          & +2.34 \\
DeepLabv3+~\citep{deeplabv3plus}& 69.36          & \textbf{82.43} & 60.88          & 64.43          & 82.77          & 56.09          & 56.68          & \textbf{88.28} & 53.81          & 60.42          & \textbf{87.72} & 56.54          & $-$7.07        & $-$4.34 \\
\bottomrule
\end{tabular}
\end{adjustbox}
\\[2pt]
{\footnotesize $^\dagger$See Appendix~\ref{app:per_class} for per-class breakdowns, background-inclusive averages, and denominator conventions.}
\label{tab:main-results-comparison}
\vspace{-4mm}
\end{table*}

\textbf{Can our SDG-pipeline generated \textsc{NavAble}\(_{\text{syn}}\) further improve performance over the baseline?}
We next evaluate synthetic augmentation under the \emph{Real+Syn} and \emph{Real+Cur+Syn} configurations (Table~\ref{tab:main-results-comparison}). Adding synthetic data to real-only training (\emph{Real+Syn} vs.\ \emph{Real}) improves five of the seven architectures: \textbf{SegMAN} gains \textbf{+7.19 mIoU} (reaching 69.39\%), \textbf{SegFormer} gains \textbf{+4.95} (62.76\%), \textbf{SAN} gains \textbf{+4.89} (66.96\%), \textbf{Mask2Former} gains \textbf{+4.18} (reaching \textbf{70.33\%}), and \textbf{EncNet} gains \textbf{+2.34} (59.58\%). Adding synthetic data on top of real+curated training (\emph{Real+Cur+Syn} vs.\ \emph{Real+Cur}) also improves five of the seven architectures: \textbf{SegMAN} achieves the highest overall score in our benchmark at \textbf{71.70\% foreground mIoU} (\textbf{+9.50 mIoU} over \emph{Real}), \textbf{SAN} reaches \textbf{68.12\%} (\textbf{+6.05} over \emph{Real}), \textbf{SegFormer} reaches \textbf{63.15\%} (\textbf{+5.34} over \emph{Real}), \textbf{EncNet} recovers by \textbf{+9.55 mIoU} over its weaker \emph{Real+Cur} baseline (48.88\% $\rightarrow$ 58.43\%), and \textbf{DeepLabv3+} edges up by +0.45 mIoU (56.09\% $\rightarrow$ 56.54\%).

In contrast, DeepLabv3+ (60.88\% $\rightarrow$ 53.81\%) and SAM~3 (70.49\% $\rightarrow$ 67.48\%) decline in aggregate mIoU relative to their \emph{Real} baselines (though SAM~3 still improves on 4 of 10 classes under synthetic augmentation, including +16.41 IoU on \texttt{door\_button} and +11.87 on \texttt{elevator}; Appendix~\ref{app:per_class}). While our experiments do not isolate a single causal mechanism, one plausible \emph{hypothesis} is differential sensitivity to synthetic appearance shifts: ImageNet-pretrained CNNs often rely strongly on local texture statistics, whereas vision transformers and global-context heads better exploit broader structural context~\citep{geirhos2019imagenet, naseer2021intriguing}. However, per-class breakdowns (Appendix~\ref{app:per_class}) show mixed behavior rather than uniform texture degradation: under \emph{Real+Syn}, DeepLabv3+ drops on \texttt{handrail} (39.79\% $\rightarrow$ 18.56\%) and \texttt{escalator} (71.23\% $\rightarrow$ 34.94\%) yet improves on \texttt{elevator} (72.90\% $\rightarrow$ 81.57\%) and \texttt{pedestrian\_signal} (32.28\% $\rightarrow$ 38.73\%), while EncNet (sharing the same ResNet-101 backbone with global context encoding) \emph{improves} on \texttt{handrail} (35.78\% $\rightarrow$ 44.14\%) and \texttt{elevator\_button} (74.34\% $\rightarrow$ 82.71\%). Thus, synthetic mixing ratios, class distributions, and architecture-specific optimization dynamics likely interact.

\begin{figure}[t]
\centering
\includegraphics[width=.67\linewidth]{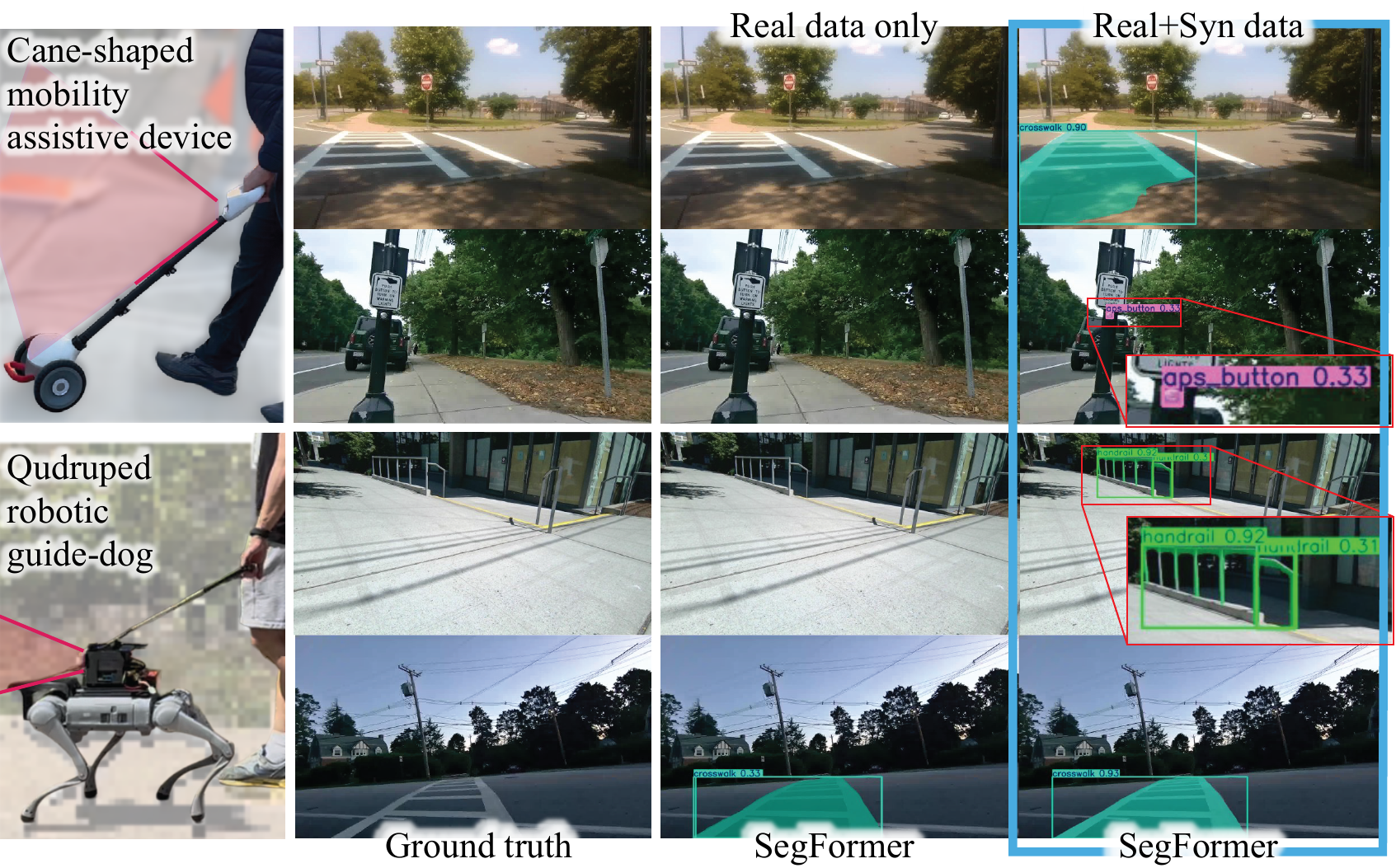}
\caption{\textbf{Embodiment evaluation.}
The first column shows the cane-shaped mobility assistive device,
Glide, and the quadruped guide-dog robot.
The remaining three columns show the corresponding input frames
and SegFormer predictions using \emph{Real} and \emph{Real+Syn}
training data, respectively.
The top two rows show Glide examples; the bottom two rows show
quadruped examples.}
\label{fig:embodiment}
\vspace{-4mm}
\end{figure}

\textbf{Per-class gains and remaining long-tail challenges.}
At a per-class level (Appendix~\ref{app:per_class}), synthetic augmentation proves especially effective for small or thin classes: under \emph{Real+Syn}, SegMAN jumps from 0.76\% to 56.75\% IoU on \texttt{pedestrian\_signal} (+55.99 IoU), SAN gains +17.78 IoU on \texttt{pedestrian\_signal} (40.14\% $\rightarrow$ 57.92\%) and +8.80 on \texttt{handrail}, SegFormer gains +11.18 IoU on \texttt{pedestrian\_signal} and +10.04 on \texttt{bus\_stop\_sign}, and Mask2Former gains +9.75 IoU on \texttt{bus\_stop\_sign} and +6.84 on \texttt{crosswalk} (reaching 91.95\%). Among the 10 evaluated foreground classes, a synthetic-augmented configuration achieves the highest IoU on 9 of 10 comparable classes for SAN and 8 of 10 for Mask2Former and SegFormer. Across all seven models, large structural categories reach high accuracy (\texttt{elevator\_button} up to 94.57\%, \texttt{crosswalk} 91.95\%, \texttt{bus\_stop} 89.70\%, and \texttt{elevator} 82.37\%), whereas small or thin long-tail classes remain harder (peaking at 52.19\% on \texttt{aps\_button}, 61.04\% on \texttt{pedestrian\_signal}, 62.80\% on \texttt{handrail}, and 69.82\% on \texttt{bus\_stop\_sign}, while hovering near 50\%--58\% on several architectures). These remaining difficulties may reflect small pixel footprints at egocentric distances. Limited geographic coverage also constrains the interpretation of these results. For example, bus stop sign images were collected predominantly on campus, leaving generalization to other signage styles and environments insufficiently evaluated. Although \textsc{NavAble} lifts \texttt{aps\_button} from 0.36\%--2.92\% IoU across zero-shot models to 52.19\% after fine-tuning, these gains may not establish reliable recognition across deployment settings. The remaining errors and coverage limitations define a measurable benchmark frontier for future research.


\textbf{Does the model improvement meaningfully transfer to real-time embodied deployment?}


Camera height, pitch, and field of view affect object visibility and projected scale. Prior work on dataset bias~\citep{torralba2011unbiased} and egocentric perception~\citep{grauman2022ego4d,zhang2024toward} motivates evaluating transfer across capture conditions. We therefore evaluate SegFormer offline on two out-of-distribution embodiment splits collected during live navigation: egocentric footage from the cane-shaped robot (157 images, 5 classes) and the quadruped robot (155 images, 9 classes), shown in Figure~\ref{fig:embodiment}. Appendix~\ref{app:viewpoint} summarizes nominal camera settings and illustrates their effects on framing. Our evaluation does not isolate viewpoint from other scene differences.

\begin{wraptable}{r}{0.62\linewidth}
\vspace{-1.0em}
\centering
\footnotesize
\setlength{\tabcolsep}{3.5pt}
\renewcommand{\arraystretch}{1.15}
\caption{\textbf{OOD embodiment evaluation} (foreground mIoU, precision, and recall) on egocentric splits from the cane-shaped and quadruped platforms. Best per platform in \textbf{bold}.}
\label{tab:embodiment}
\begin{tabular}{l ccc ccc}
\toprule
& \multicolumn{3}{c}{\textbf{Cane-shaped}} & \multicolumn{3}{c}{\textbf{Quadruped}} \\
\cmidrule(lr){2-4}\cmidrule(lr){5-7}
\textbf{Model} & mIoU & mPrec & mRec & mIoU & mPrec & mRec \\
\midrule
SAN-ZS          & 10.27 & 10.63 & \textbf{45.34} & \phantom{0}9.95 & 10.78 & 32.51 \\
SegFormer (R)   & 11.94 & 24.69 & 39.93          & 21.56           & 30.73 & 38.90 \\
SegFormer (R+S) & \textbf{16.17} & \textbf{31.44} & 38.22 & \textbf{23.08} & \textbf{34.24} & \textbf{39.25} \\
\bottomrule
\end{tabular}
\vspace{-0.5em}
\end{wraptable}

As shown in Table~\ref{tab:embodiment}, fine-tuning on \textsc{NavAble}-Real exceeds SAN zero-shot on both platforms (10.27\% $\rightarrow$ 11.94\% on the cane-shaped robot; 9.95\% $\rightarrow$ 21.56\% on the quadruped), and adding synthetic supervision (\emph{Real+Syn}) yields further relative gains to \textbf{16.17\%} (+4.23 mIoU) and \textbf{23.08\%} (+1.52 mIoU), with per-class improvements of +9.64 IoU on \texttt{crosswalk} (cane), +8.50 on \texttt{aps\_button} (quadruped), +7.36 on \texttt{pedestrian\_signal} (quadruped), and +6.63 on \texttt{handrail} (quadruped; Appendix~\ref{app:embodiment_per_class}). We emphasize that this improvement is strictly \emph{relative}: while \textsc{NavAble}\(_{\text{syn}}\) transfers in the right direction across two structurally distinct platforms without viewpoint retargeting, absolute accuracy of 16.17\%--23.08\% mIoU does not establish deployment readiness for a real assistive device. Closing this remaining embodiment gap---for example, by using our SDG pipeline to render platform-matched camera heights and pitch angles---is a primary direction motivated by these results.


\section{Concluding remarks}
\label{sec:conclusion}
We introduced \textsc{NavAble}, a large-scale dataset and synthetic data generation pipeline for accessibility-critical perception in BLV navigation. By combining curated open-source data, newly collected real-world annotations, and controllable synthetic rendering in Isaac~Sim, \textsc{NavAble} directly addresses severe data scarcity and annotation inconsistency in this domain. While state-of-the-art open-vocabulary models struggle in zero-shot settings (at most 26.07\% foreground mIoU), fine-tuning on \textsc{NavAble} reaches up to 71.70\% foreground mIoU, and adding \textsc{NavAble}$_{syn}$ improves five of the seven evaluated architectures by up to +7.19 foreground mIoU (+9.50 with curated data) while transferring positively to out-of-distribution assistive robot viewpoints. We release the dataset, 3D asset library, Isaac~Sim extension, and generation pipeline to support future research in accessible robotics and embodied navigation.

\textbf{Limitations and Future Work:} Although \textsc{NavAble} substantially expands the coverage of accessibility-critical classes, many important BLV-relevant objects and environmental cues remain outside the current taxonomy. In addition, despite our efforts to include diverse environments and viewpoints, the geographic diversity of our real-world data remains limited, motivating broader data collection in future iterations. Portions of the synthetic pipeline still require human intervention, including object placement and trajectory recording. Additional revolving-door evaluation on 100 target-positive frames from six video sequences reveals limited transfer for several architectures when positive real training examples of a class are absent (see Appendix~\ref{app:revolving_door}). Finally, although our synthetic data improves transfer robustness, real-world deployment of the model in combination with assistive mobile robot navigation systems remains an open challenge.

Looking forward, recent advances in foundation models for image understanding, language-guided scene generation, and text-to-3D asset synthesis present opportunities to further automate and scale accessibility-oriented simulation pipelines. Future work could also explore complementary audio, such as audible pedestrian signals, while addressing urban noise and source localization. We hope \textsc{NavAble} motivates broader research at the intersection of embodied AI and accessibility, ultimately enabling perception systems that better support safe and independent mobility for BLV individuals.

\section*{Acknowledgments}
This work is supported by NIH (R21EY037411), NSF (2427788, 2541914, 2529277, and IIS-2152077), the Electronics and Telecommunications Research Institute (ETRI), and the NVIDIA Academic Grant Program. We also thank Glidance for sharing data used in our evaluation.

\newpage
\bibliographystyle{unsrtnat} 
\bibliography{references}

\newpage
\appendix

\section{Detailed training settings}
\label{app:training_details}

This appendix complements Sec.~\ref{subsec:setup} with the shared
training recipe, per-architecture hyperparameters, per-configuration
iteration budgets, architecture-specific deviations, and training
hardware. Throughout this appendix, the four training-data
configurations are abbreviated as \emph{Real}, \emph{Real+Cur},
\emph{Real+Syn}, and \emph{Real+Cur+Syn}, where
\emph{Real} / \emph{Cur} / \emph{Syn} denote the \textsc{NavAble}-Real,
\textsc{NavAble}-Curation, and \textsc{NavAble}-Synthetic subsets,
respectively.

\paragraph{Shared training recipe.}
The five original architectures (DeepLabv3+, EncNet, SegFormer, Mask2Former, and SAN) share a common augmentation pipeline (random resize jittering with long-side scale ratio $0.5{\times}$--$2.0{\times}$, category-balanced random cropping with \texttt{cat\_max\_ratio}$=0.75$, horizontal flipping with $p{=}0.5$, and photometric distortion) and a linear-warmup followed by a polynomial-decay learning-rate schedule, while otherwise following each model's standard recipe. We apply per-class loss weighting with the background class down-weighted to $0.10$ and all $11$ foreground classes set to $1.00$. Within each architecture, the four data configurations share identical augmentation, optimizer, and schedule settings---only the training corpus and iteration budget differ---so that any change in performance is attributable to the data. Iteration budgets are scaled with corpus size to keep approximate epoch coverage roughly constant, and the final checkpoint per run is selected by best foreground mIoU on the held-out \textsc{NavAble}-Real validation split.

\paragraph{Per-architecture base settings.} Table~\ref{tab:training_settings} (top) lists the backbone, input crop, optimizer (with per-component learning-rate multipliers), weight decay, batch size, and AMP usage of each architecture. Within each architecture, these values are held fixed across all four training-data configurations.

\paragraph{Iteration budgets across data configurations.} Table~\ref{tab:training_settings} (bottom) reports the iteration budget for each (architecture, configuration) pair. Budgets are chosen so that the approximate number of epochs over the training corpus is comparable across the four configurations within an architecture, ensuring that any change in test performance reflects the training data rather than the optimization budget. For DeepLabv3+, EncNet, and SegFormer, the per-GPU batch size and the number of GPUs are held fixed across all configurations, so iteration budgets are determined solely by corpus size and are identical across these three architectures. For each run, the final checkpoint is selected by best foreground mIoU on the held-out \textsc{NavAble}-Real validation split.

\paragraph{Mask2Former-specific deviations.} Mask2Former is the one architectural outlier in our setup, and we deviate from its default recipe in two ways. \textit{Numerical stability.} The Hungarian matcher emits NaN cost matrices under fp16, so we disable AMP (denoted as $\dagger$ in Table~\ref{tab:training_settings}), train in full precision with gradient clipping at max-norm $0.1$, and apply a backbone learning-rate multiplier of $0.1{\times}$. \textit{Memory and effective-batch scaling.} Swin-L at $640{\times}640$ exceeds memory on a subset of our cluster nodes, so we adjust the per-GPU batch size and the number of GPUs per configuration, and rescale iteration counts so that the total samples seen during training ($\text{iters}\times\text{effective batch}$) remain comparable to the non-M2F runs on the same corpus -- which is why the Mask2Former column in Table~\ref{tab:training_settings} (bottom) differs from the other three CNN/transformer architectures. 

\paragraph{Recent baseline schedules (SegMAN and SAM~3).}
To evaluate recent segmentation architectures under identical data splits (Table~\ref{tab:main-results-comparison}), we fine-tune SegMAN~\citep{segman} and SAM~3~\citep{sam3} across all four training-data configurations (\emph{Real}, \emph{Real+Cur}, \emph{Real+Syn}, and \emph{Real+Cur+Syn}) for a 20-epoch schedule. For SegMAN, checkpoints are selected by best validation mIoU (epochs 20, 18, 20, and 19 across \emph{Real}, \emph{Real+Cur}, \emph{Real+Syn}, and \emph{Real+Cur+Syn}). For SAM~3, multi-source runs (\emph{Real+Cur}, \emph{Real+Syn}, \emph{Real+Cur+Syn}) use a source-aware label policy in which partial-label curated and synthetic images are queried only for their annotated classes while densely labeled \emph{Real} images are queried across all classes, evaluated at the final 20-epoch checkpoint (which is also the validation-best AP checkpoint on \emph{Real}). Separately, in preliminary 10-epoch runs, SAM~3 recorded 50.44, 37.83, 48.69, and 49.65 mIoU and SegMAN recorded 60.92, 61.21, 58.27, and 60.03 mIoU across \emph{Real}, \emph{Real+Cur}, \emph{Real+Syn}, and \emph{Real+Cur+Syn}, respectively.

\paragraph{Embodiment-viewpoint experiments.} For the out-of-distribution embodiment evaluation on the cane-shaped and quadruped platforms (Sec.~\ref{sec:experiments}, Table~\ref{tab:embodiment}), we evaluate the SegFormer (MiT-B5) checkpoints trained under the \emph{Real} and \emph{Real+Syn} configurations from Table~\ref{tab:training_settings} without any platform-specific fine-tuning or hyperparameter changes.

\paragraph{Hardware.} All models are trained on a combination of two compute resources: a university GPU cluster equipped with $2$--$4{\times}$ NVIDIA L40S ($48$\,GB) nodes, and a local lab workstation with NVIDIA RTX PRO 6000 Blackwell Max-Q GPUs.

\begin{table}[t]
\centering
\small
\caption{Per-architecture training settings for \textsc{NavAble} fine-tuning. Configurations \emph{Real}, \emph{Real+Cur}, \emph{Real+Syn}, and \emph{Real+Cur+Syn} combine the \textsc{NavAble}-Real (\emph{Real}), \textsc{NavAble}-Curation (\emph{Cur}), and \textsc{NavAble}-Synthetic (\emph{Syn}) subsets as defined in Sec.~\ref{subsec:setup}. \textbf{(Top)} Base settings per architecture. \textbf{(Bottom)} Iteration budget per (architecture, configuration) pair.}
\label{tab:training_settings}

\centerline{%
\begin{tabular}{l l c l c c c}
\toprule
\textbf{Model} & \textbf{Backbone} & \textbf{Crop} & \textbf{Optimizer (base lr)} & \textbf{WD} & \textbf{Batch} & \textbf{AMP} \\
\midrule
DeepLabv3+   & R-101-d8       & 512 & SGD\ \ ($5{\times}10^{-3}$, mom.\ $0.9$)        & $5{\times}10^{-4}$ & $8$        & \checkmark \\
EncNet       & R-101-d8       & 512 & SGD\ \ ($5{\times}10^{-3}$, mom.\ $0.9$)        & $5{\times}10^{-4}$ & $8$        & \checkmark \\
SegFormer    & MiT-B5         & 640 & AdamW ($1{\times}10^{-5}$, head $10\times$)     & $0.01$             & $8$        & \checkmark \\
Mask2Former  & Swin-L         & 640 & AdamW (variable, backbone $0.1\times$)          & $0.05$             & $4$--$8$   & $\dagger$  \\
SAN          & CLIP ViT-B/16  & 640 & AdamW ($1{\times}10^{-4}$, encoder $0.1\times$) & $1{\times}10^{-4}$ & $24$       & \checkmark \\
\bottomrule
\end{tabular}%
}
\par\smallskip

\centerline{%
\begin{tabular}{l c c c c c}
\toprule
\textbf{Config} & \textbf{DeepLabv3+} & \textbf{EncNet} & \textbf{SegFormer} & \textbf{Mask2Former} & \textbf{SAN} \\
\midrule
Real                  & 16K  & 16K  & 16K  & 32K  & 6K  \\
Real+Cur                & 80K  & 80K  & 80K  & 160K & 30K \\
Real+Syn                & 220K & 220K & 220K & 220K & 180K \\
Real+Cur+Syn              & 324K & 324K & 324K & 324K & 324K \\
\bottomrule
\end{tabular}%
}

\end{table}

\section{Annotation Protocol, Verification, and Data Filtering}
\label{app:annotation_filtering}

This appendix details the annotation and quality-control procedures for the newly collected real-world images and the manual filtering of public real-image sources. Both components followed a shared class taxonomy and consistent mask conventions.

\begin{figure}[htbp]
\centering
\includegraphics[width=0.99\linewidth]{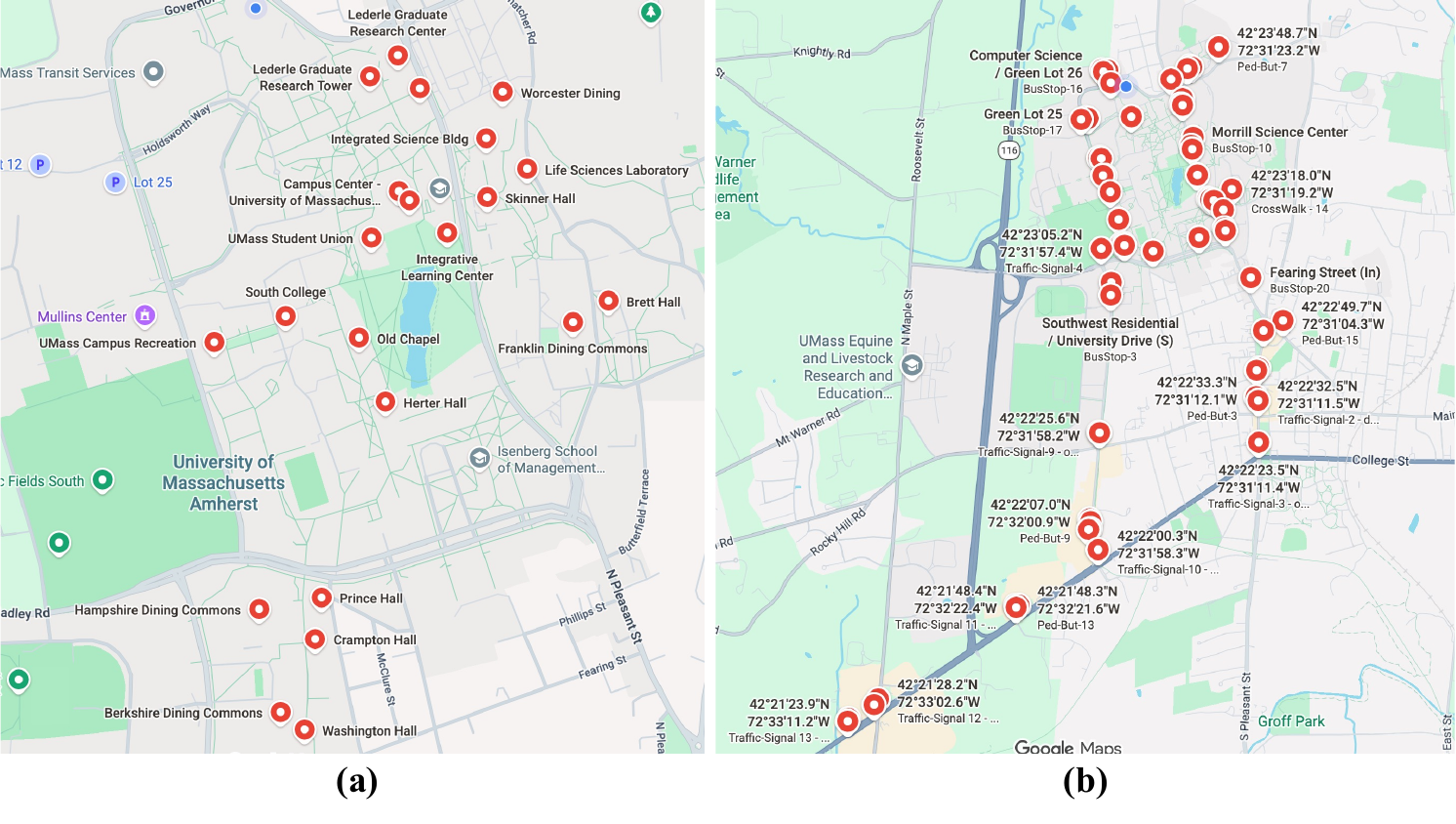}
\caption{\textbf{Real-World Data Collection Coverage}. (a) Indoor collection sites on the University of Massachusetts Amherst campus.
(b) Outdoor sites across campus and surrounding Amherst, Massachusetts.
Maps from Google Maps.}
\label{fig:coverage}
\end{figure}

\paragraph{Collection and initial screening.}
We recorded video trajectories at $30$\,fps using a chest-mounted Intel RealSense D435 across $50$ indoor and outdoor locations, yielding approximately $153$K frames. We subsampled every fifteenth frame to reduce near-duplicate content. Two annotators then screened the sampled images using a predefined rubric, excluding motion or defocus blur, underexposure or overexposure, and corrupted content. This process produced approximately $6.8$K candidate images for annotation. Figure~\ref{fig:coverage} shows the geographic coverage of
indoor and outdoor collection sites.

\paragraph{Annotation team and shared rules.}
Four annotators worked on a shared Label Studio server~\cite{Labelstudio}, with one compiler overseeing consistency across annotators and review rounds. The team included student researchers in accessibility and computer vision. Before annotation, in-person sessions established the following rules:
\begin{enumerate}
    \item \textbf{Consistent labels and class indices.} Annotators used a fixed class index, label vocabulary, and shared mask standards.
    \item \textbf{Multi-class completeness.} Every visible instance of a target category was annotated, rather than a selected subset of objects.
    \item \textbf{Class-specific mask definitions.} Mask conventions were agreed in advance, including the treatment of round versus rectangular elevator buttons and the representation of curved objects. Because our taxonomy is object-centric rather than traversability-centric, only visible object pixels are annotated; for instance, if a crosswalk is partially blocked by a parked vehicle or construction barrier, only the visible crosswalk pixels are labeled while the out-of-taxonomy obstacle remains background.
    \item \textbf{Consistent export format.} Annotations were exported as COCO-format polygons for downstream training and evaluation.
    \item \textbf{Consistency across data sources.} Collected, curated, and synthetic annotations followed the same mask definitions. Public annotations incompatible with these definitions were excluded during curation.
\end{enumerate}

\paragraph{Three-pass verification and disagreement handling.}
Each annotator worked on a fixed image set, and annotations underwent three full passes with cross-checking and iterative correction. Annotation sets were first cross-checked by other annotators for missed instances and returned with comments. Revised annotations were reviewed by the compiler against the agreed class and mask definitions. Images for which the annotator and compiler could not reach consensus were excluded rather than retained through a unilateral decision. Such cases included uncertainty about partially occluded elevator buttons and disagreement over whether a button's rectangular housing should be included in its mask. This process excluded approximately $1.3$K of the $6.8$K candidates ($19\%$), leaving $5{,}581$ retained images across the 10 primary benchmark classes ($3{,}703$ train, $396$ validation, and $1{,}482$ test images; $5{,}681$ total collected records when including the $100$ supplementary \texttt{revolving\_door} frames evaluated in Appendix~\ref{app:revolving_door}).

\paragraph{Independent reannotation agreement.}
Two annotators independently reannotated a held-out subset of $100$ images. Comparison of their annotations yielded a pairwise mask mIoU of $97.2\%$. This independent reannotation audit complemented the multi-pass verification applied throughout the collected dataset.

\paragraph{Public-data inclusion and exclusion.}
Curation combined seven source-level exclusion criteria before download (defined in the released dataset card) with exhaustive manual review of the downloaded image pool. Every image was individually inspected with its segmentation mask overlaid. Inclusion required recognizable photographic content and annotations consistent with the shared class and mask definitions. Images were rejected for any of the following six reasons:
\begin{enumerate}
    \item Mask boundaries were misaligned with the depicted object.
    \item The annotation label did not match the object class.
    \item A visible target instance was left unannotated.
    \item The annotation was degenerate.
    \item The target object was not recognizable.
    \item The image contained non-photographic content.
\end{enumerate}
This review addressed the inconsistent boundaries, missing instances, and label noise illustrated in Figure~\ref{fig:data-curation}.

\paragraph{Filtering statistics.}
The initial public real-image pool contained $45{,}060$ images before image-level filtering. Manual review removed $9{,}235$ images ($20.5\%$), leaving $35{,}825$ images ($\sim$36K; $36{,}466$ total unique curated images in the final released \textsc{NavAble}\(_{\text{cur}}\) split after incorporating $641$ additional curated samples). The removal rates for the reported training, validation, and test splits of the initial pool were $23.5\%$, $8.9\%$, and $16.1\%$, respectively. Table~\ref{tab:annotation_filtering_statistics} summarizes these filtering statistics, and Table~\ref{tab:per_class_counts} reports the complete per-class image and instance composition across all three data sources corresponding to Figure~\ref{fig:overview}.

\begin{table}[htbp]
\centering
\small
\caption{Manual filtering statistics for the initial public real-image pool. Aggregate image counts and split-specific removal rates are reported separately.}
\label{tab:annotation_filtering_statistics}
\begin{tabular}{lr}
\toprule
\textbf{Curation statistic} & \textbf{Value} \\
\midrule
Images before filtering & $45{,}060$ \\
Images excluded & $9{,}235$ \\
Images retained (initial pool) & $35{,}825$ \\
\midrule
Overall removal rate & $20.5\%$ \\
Training-split removal rate & $23.5\%$ \\
Validation-split removal rate & $8.9\%$ \\
Test-split removal rate & $16.1\%$ \\
\bottomrule
\end{tabular}
\end{table}

\begin{table}[htbp]
\centering
\footnotesize
\caption{Per-class composition of \textsc{NavAble} corresponding to Figure~\ref{fig:overview} and the Hugging Face release. Because multi-class images contain multiple target categories, per-class image counts sum to more than the unique image total ($41{,}345$ class-appearances across $36{,}466$ unique curated open-source images; $7{,}956$ class-appearances across $5{,}581$ primary collected real images, plus $100$ supplementary \texttt{revolving\_door} frames totaling $5{,}681$ collected real records; and $451{,}100$ class-appearances across $450{,}860$ unique synthetic frames containing $781$K total instance annotations).}
\label{tab:per_class_counts}
\begin{tabular}{lrrr}
\toprule
\textbf{Class} & \textbf{Curated Open-Source} & \textbf{Collected Real} & \textbf{Synthetic} \\
               & \textbf{(imgs / instances)}  & \textbf{(imgs / instances)} & \textbf{(imgs)} \\
\midrule
elevator           &        15 /         15 & 1{,}663 / 1{,}844 & 23{,}400 \\
elevator\_button   & 4{,}401 / 39{,}033     & 1{,}945 / 1{,}952 & 23{,}400 \\
door\_button       &         0 /          0 & 2{,}122 / 2{,}426 & 45{,}400 \\
crosswalk          & 27{,}786 / 49{,}464    &    434 /    700   & 54{,}000 \\
pedestrian\_signal & 6{,}650 / 19{,}407     &    456 /    627   & 44{,}700 \\
aps\_button        &         0 /          0 &    295 /    301   & 62{,}700 \\
bus\_stop          &         0 /          0 &    435 /    469   & 60{,}800 \\
bus\_stop\_sign    &         0 /          0 &    210 /    291   & 60{,}500 \\
revolving\_door    &         0 /          0 & 100$^\dagger$ / 100$^\dagger$ & 25{,}000 \\
handrail           & 1{,}197 / 3{,}843      &    206 /    753   & 44{,}800 \\
escalator          & 1{,}296 / 1{,}684      &    190 /    242   &  6{,}400 \\
\midrule
\textbf{Total (unique)} & \textbf{36{,}466 / 113{,}446} & \textbf{5{,}681$^\dagger$ / 9{,}705} & \textbf{450{,}860 (781K inst.)} \\
\bottomrule
\end{tabular}
\\[2pt]
{\footnotesize $^\dagger$Comprises $5{,}581$ primary collected real-world images across 10 classes ($3{,}703$ train / $396$ val / $1{,}482$ test; $9{,}605$ instances) and $100$ supplementary \texttt{revolving\_door} frames evaluated in Appendix~\ref{app:revolving_door}.}
\end{table}

\clearpage
\section{Details of expanded \textsc{NavAble}$_{syn}$ and manual effort analysis}
\label{app:synth_details}

\paragraph{Lighting variation and compositional rendering.}
Lighting variation is incorporated as part of the environment axis. Each scene is exported from Unreal to USDZ with baked textures, preserving the original lighting fidelity. We then compose additional lighting conditions (e.g., Day, Sunset, Night outdoors; ambient and directional variants indoors) as separate USD layers that can be toggled at render time. This design enables paired observations of the same trajectory under different illumination without additional data collection.

\paragraph{Synthetic-real distribution gap.}
We measured a pooled synthetic-real Fr\'echet Inception Distance (FID) of 157.3 across the 11 categories. For comparison, object-framed and scene-framed images within the real data yielded an FID of 132.6, suggesting that framing and composition also influence this comparison. These results indicate a remaining feature-space distribution gap. They do not establish photorealism or closure of the sim-to-real gap. We therefore evaluate the utility of synthetic data primarily through downstream segmentation performance on real images.

\begin{figure}[h!]
\centering
\includegraphics[width=0.85\linewidth]{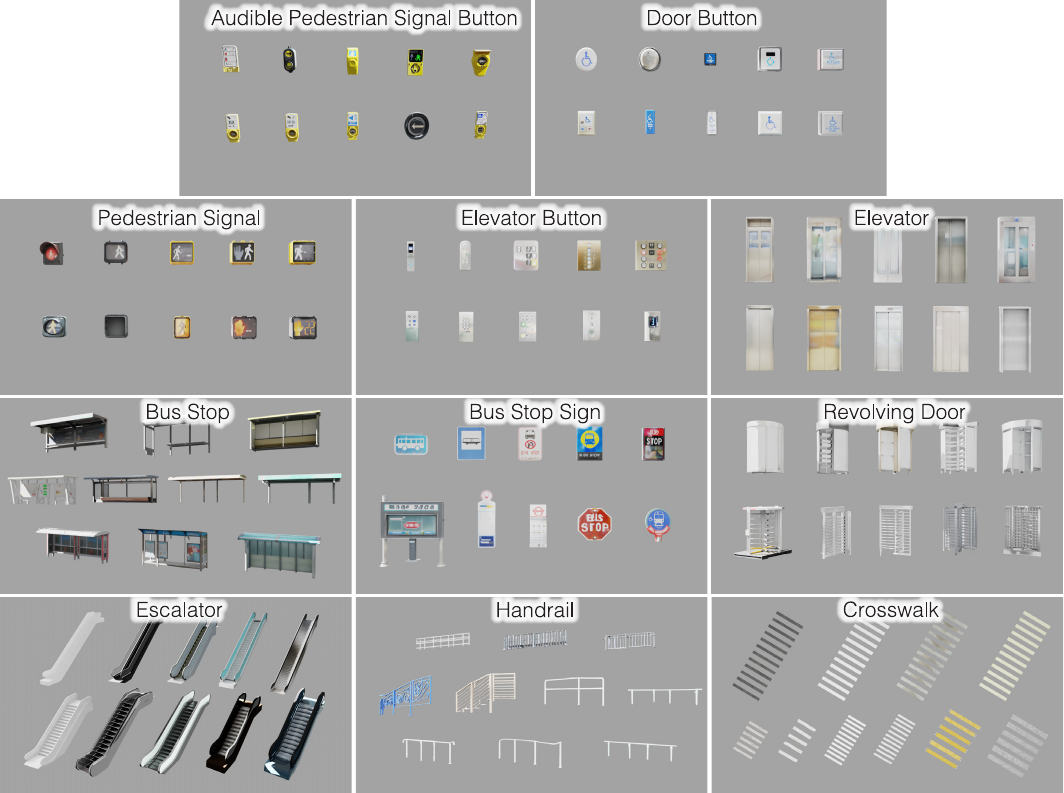}
\caption{Samples of our synthetic 3D assets. \textsc{NavAble} includes synthetic assets of 15 APS buttons, 91 door buttons, 66 pedestrian signals, 74 elevator buttons, 111 elevators, 61 bus stops, 10 bus stop signs, 48 revolving doors, 5 escalators, 29 handrails, and 10 crosswalks. Except for crosswalks and escalators, all assets are generated through our synthetic data pipeline.}
\label{fig:append_assets}
\end{figure}

\paragraph{Stage-wise asset yield and manual intervention cost.}
To quantify the human effort required by the \textsc{NavAble} Generator (visualized in Figure~\ref{fig:append_assets}), we measured per-stage throughput and rejection rates on a 150-image pilot sample across 3 of the 11 target classes: \texttt{aps\_button} (48 crawled images), \texttt{escalator} (50), and \texttt{pedestrian\_signal} (50), totaling 148 successfully crawled images from 150 queries (Table~\ref{tab:synth_yield}). Candidate screening prior to 3D reconstruction is fully automated: Gemini VLM validation removed 105 of the 148 crawled images (primarily non-photographic diagrams/renders at $\sim$32\%, stock watermarks at $\sim$22\%, wrong object classes at $\sim$22\%, collages/slides at $\sim$10\%, truncated framing at $\sim$10\%, and background cutouts/hand occlusions at $\sim$4\%), and automated mask-quality filtering removed an additional 9 of the 43 remaining images. Human intervention in the asset pipeline is thus limited to inspecting the 34 reconstructed USDZ meshes, which takes approximately 1~minute per 10~meshes and identified 10 geometry or texture failures (4 escalators modeled as planar inclines without step geometry, 2 escalator meshes unrelated to the input image, 3 meshes with missing textures, and 1 APS button with poor morphology), yielding 24 simulation-ready assets (16.0\% end-to-end yield from initial queries, or 22.9\% for \texttt{aps\_button}, 10.0\% for \texttt{escalator}, and 16.0\% for \texttt{pedestrian\_signal}). During scene composition and rendering, 10--20~minutes of interactive placement and gamepad trajectory teleoperation replaces 5--6~hours of equivalent manual capture and annotation. Finally, automated post-render filtering removes frames with empty foreground masks ($<$1\% of generated frames), followed by a lightweight human quality check requiring approximately 1~minute per 100~rendered images.

\begin{table}[htbp]
\centering
\small
\caption{Stage-wise filtering yield and human vs.\ automated screening scope on a 150-query pilot across three classes (\texttt{aps\_button}, \texttt{escalator}, and \texttt{pedestrian\_signal}).}
\label{tab:synth_yield}
\begin{tabular}{l l rrr r}
\toprule
\textbf{Stage} & \textbf{Screening Mode} & \textbf{In} & \textbf{Passed} & \textbf{Dropped} & \textbf{Stage Yield} \\
\midrule
Web crawl              & Automated           & 150 & 148 &   2 &  98.7\% \\
VLM validation (Gemini)& Automated           & 148 &  43 & 105 &  29.1\% \\
Mask quality check     & Automated           &  43 &  34 &   9 &  79.1\% \\
3D reconstruction (SAM~3D) & Automated       &  34 &  34 &   0 & 100.0\% \\
USDZ conversion        & Automated           &  34 &  34 &   0 & 100.0\% \\
Mesh quality review    & Manual ($\sim$1 min / 10 meshes) & 34 & 24 & 10 & 70.6\% \\
\midrule
\textbf{End-to-end total} & --               & \textbf{150} & \textbf{24} & \textbf{126} & \textbf{16.0\%} \\
\bottomrule
\end{tabular}
\end{table}

\clearpage

\section{Per-class evaluation results}
\label{app:per_class}

Tables~\ref{tab:perclass-segman}--\ref{tab:perclass-encnet} report per-class IoU, 10-class foreground-only mean IoU (\textbf{mean 10 FG classes}), and 11-class background-inclusive mean IoU (\textbf{mean 11 classes w/ bg}) on the held-out \textsc{NavAble}-Real test split (1{,}482 images) for all seven fine-tuned architectures under the four training-data configurations described in Sec.~\ref{subsec:setup}, while Table~\ref{tab:perclass-zs} reports per-class IoU for the open-vocabulary zero-shot baselines on the same split. Because the 1{,}482-image split contains the 10 foreground classes present in \emph{Real} (with \texttt{revolving\_door} evaluated separately on 100 target-positive frames in Appendix~\ref{app:revolving_door}), all tables below list those 10 evaluated foreground categories so all models share an identical 10-class foreground denominator. For the three runs that emit false-positive \texttt{revolving\_door} predictions on the 1{,}482-image split (DeepLabv3+ \emph{+Syn} and EncNet \emph{+Syn}/\emph{+Cur+Syn}, which receive $\text{IoU}=0$ on that class), the parenthesized values in the mean rows also report the 11-class foreground and 12-class overall averages when $\text{IoU}=0$ is included in the denominator. Bold marks the best configuration per class within each architecture.

As shown in Table~\ref{tab:perclass-zs}, zero-shot failure is structured rather than uniform: open-vocabulary models are competitive only on large, web-common structural categories (\texttt{elevator} up to 67.63\% and \texttt{bus\_stop} up to 63.50\% IoU), but collapse on the small functional components that motivate \textsc{NavAble}---\texttt{aps\_button} scores 0.36\%--1.46\% IoU across all four added models, \texttt{handrail} scores 1.09\%--15.05\%, and \texttt{bus\_stop\_sign} scores 2.83\%--13.80\%, while \texttt{elevator\_button} remains at 3.07\%--43.64\% despite substantial test support.

\begin{table}[h!]
\centering
\small
\caption{Per-class zero-shot IoU (\%) for open-vocabulary baselines on the held-out \textsc{NavAble}-Real test split (1{,}482 images, 10 present foreground classes; see Appendix~\ref{app:revolving_door} for \texttt{revolving\_door}). SAN-ZS scores 14.94\% mIoU on the same split. Best model per class in bold.}
\label{tab:perclass-zs}
\setlength{\tabcolsep}{6pt}
\begin{tabular}{l rrrr}
\toprule
\textbf{Class} & \textbf{CAT-Seg-L} & \textbf{SED-L} & \textbf{SAM~3} & \textbf{YOLOE-M} \\
\midrule
elevator           & \textbf{67.63} & 51.88          & 47.27          & 19.26 \\
elevator\_button   & 20.67          & \textbf{43.64} & 3.07           & 17.68 \\
door\_button       & 10.82          & 8.62           & \textbf{16.34} & 0.66 \\
crosswalk          & \textbf{21.39} & 19.03          & 11.20          & 14.42 \\
pedestrian\_signal & \textbf{20.68} & 18.52          & 1.31           & 4.78 \\
aps\_button        & 1.43           & 0.36           & 0.51           & \textbf{1.46} \\
bus\_stop          & \textbf{63.50} & 61.38          & 60.47          & 46.26 \\
bus\_stop\_sign    & \textbf{13.80} & 2.83           & 9.90           & 3.95 \\
handrail           & 4.34           & 1.09           & \textbf{15.05} & 4.74 \\
escalator          & \textbf{36.45} & 33.33          & 25.50          & 36.17 \\
\midrule
\textbf{mIoU (10 FG classes)} & \textbf{26.07} & 24.07 & 19.06       & 14.94 \\
\bottomrule
\end{tabular}
\end{table}

\begin{table}[ht]
\centering
\small
\caption{Per-class IoU (\%) for \textbf{SegMAN} (20-epoch schedule) on the held-out \textsc{NavAble}-Real test split ($N{=}1{,}482$). Best configuration per class in bold.}
\label{tab:perclass-segman}
\setlength{\tabcolsep}{5pt}
\begin{tabular}{l rrrr}
\toprule
\textbf{Class} & \textbf{Real} & \textbf{+Cur} & \textbf{+Syn} & \textbf{+Cur+Syn} \\
\midrule
background         & 93.31 & 93.83          & 93.25          & \textbf{94.40} \\
elevator           & 60.86 & 65.06          & 58.59          & \textbf{66.82} \\
elevator\_button   & 91.89 & 92.42          & 93.28          & \textbf{94.57} \\
door\_button       & 77.97 & 78.39          & 80.05          & \textbf{81.93} \\
crosswalk          & 88.52 & 90.50          & 90.48          & \textbf{91.73} \\
pedestrian\_signal &  0.76 & 57.51          & 56.75          & \textbf{57.91} \\
aps\_button        & 43.20 & 37.09          & \textbf{46.95} & 46.51 \\
bus\_stop          & 85.26 & 81.64          & 88.31          & \textbf{89.70} \\
bus\_stop\_sign    & 45.35 & \textbf{55.27} & 45.20          & 53.99 \\
handrail           & 52.48 & \textbf{62.80} & 57.94          & 60.48 \\
escalator          & 75.73 & \textbf{77.17} & 76.33          & 73.38 \\
\midrule
\textbf{mean (10 FG classes)}    & 62.20 & 69.78 & 69.39 & \textbf{71.70} \\
\textbf{mean (11 classes w/ bg)} & 65.03 & 71.97 & 71.56 & \textbf{73.77} \\
\bottomrule
\end{tabular}
\end{table}

\begin{table}[ht]
\centering
\small
\caption{Per-class IoU (\%) for \textbf{SAM~3} (20-epoch schedule) on the held-out \textsc{NavAble}-Real test split ($N{=}1{,}482$). Best configuration per class in bold.}
\label{tab:perclass-sam3}
\setlength{\tabcolsep}{5pt}
\begin{tabular}{l rrrr}
\toprule
\textbf{Class} & \textbf{Real} & \textbf{+Cur} & \textbf{+Syn} & \textbf{+Cur+Syn} \\
\midrule
background         & 94.10          & 94.76 & 95.28          & \textbf{95.70} \\
elevator           & 70.02          & 76.04 & 79.04          & \textbf{81.89} \\
elevator\_button   & \textbf{92.96} & 87.69 & 92.35          & 90.84 \\
door\_button       & 60.82          & 72.14 & 67.70          & \textbf{77.23} \\
crosswalk          & \textbf{83.47} & 77.10 & 75.48          & 71.93 \\
pedestrian\_signal & 60.18          & 60.09 & \textbf{61.04} & 54.90 \\
aps\_button        & \textbf{45.69} & 41.76 & 38.68          & 34.18 \\
bus\_stop          & 87.94          & 86.79 & \textbf{89.30} & 88.85 \\
bus\_stop\_sign    & \textbf{69.82} & 62.66 & 47.99          & 41.33 \\
handrail           & \textbf{59.49} & 51.01 & 57.85          & 54.12 \\
escalator          & \textbf{74.52} & 58.97 & 65.41          & 58.18 \\
\midrule
\textbf{mean (10 FG classes)}    & \textbf{70.49} & 67.42 & 67.48 & 65.35 \\
\textbf{mean (11 classes w/ bg)} & \textbf{72.64} & 69.91 & 70.01 & 68.11 \\
\bottomrule
\end{tabular}
\end{table}

\begin{table}[ht]
\centering
\small
\caption{Per-class IoU (\%) for \textbf{Mask2Former} (Swin-L) on the held-out \textsc{NavAble}-Real test split ($N{=}1{,}482$). Best configuration per class in bold.}
\label{tab:perclass-m2f}
\setlength{\tabcolsep}{5pt}
\begin{tabular}{l rrrr}
\toprule
\textbf{Class} & \textbf{Real} & \textbf{+Cur} & \textbf{+Syn} & \textbf{+Cur+Syn} \\
\midrule
background         & 95.76          & 95.73          & \textbf{96.14} & 95.73 \\
elevator           & \textbf{82.37} & 81.78          & 81.89          & 81.69 \\
elevator\_button   & 91.19          & 89.82          & \textbf{92.83} & 91.51 \\
door\_button       & 76.34          & 74.20          & \textbf{77.99} & 75.91 \\
crosswalk          & 85.11          & 87.39          & \textbf{91.95} & 87.37 \\
pedestrian\_signal & 42.89          & 56.30          & 51.70          & \textbf{56.98} \\
aps\_button        & 42.54          & 43.77          & 43.08          & \textbf{46.13} \\
bus\_stop          & 82.25          & 82.07          & \textbf{86.85} & 85.12 \\
bus\_stop\_sign    & 40.23          & 46.59          & \textbf{49.98} & 36.41 \\
handrail           & 48.05          & 52.47          & 53.11          & \textbf{53.12} \\
escalator          & 70.50          & \textbf{74.07} & 73.87          & 68.33 \\
\midrule
\textbf{mean (10 FG classes)}    & 66.15 & 68.85 & \textbf{70.33} & 68.26 \\
\textbf{mean (11 classes w/ bg)} & 68.84 & 71.29 & \textbf{72.67} & 70.75 \\
\bottomrule
\end{tabular}
\end{table}

\begin{table}[ht]
\centering
\small
\caption{Per-class IoU (\%) for \textbf{SAN} (CLIP ViT-B/16). Best comparable configuration per class in bold.}
\label{tab:perclass-san}
\setlength{\tabcolsep}{5pt}
\begin{tabular}{l rrrr}
\toprule
\textbf{Class} & \textbf{Real} & \textbf{+Cur} & \textbf{+Syn} & \textbf{+Cur+Syn}$^{\dagger}$ \\
\midrule
elevator           & 78.58 & \textbf{79.38} & 68.83          & 81.52 \\
elevator\_button   & 88.73 & 87.18          & \textbf{92.90} & 83.01 \\
door\_button       & 68.03 & 66.29          & \textbf{75.26} & 69.09 \\
crosswalk          & 90.72 & 84.06          & \textbf{91.60} & 90.63 \\
pedestrian\_signal & 40.14 & 49.01          & \textbf{57.92} & 48.38 \\
aps\_button        & 46.92 & 48.27          & \textbf{52.19} & 49.40 \\
bus\_stop          & 82.19 & 79.26          & \textbf{85.03} & 81.50 \\
bus\_stop\_sign    & 22.87 & 27.36          & \textbf{29.19} & 25.32 \\
handrail           & 41.98 & 40.77          & \textbf{50.78} & 46.46 \\
escalator          & 60.56 & 63.62          & \textbf{65.94} & 63.44 \\
\midrule
\textbf{mean (10 FG classes)} & 62.07 & 62.52 & \textbf{66.96} & 63.88$^{\dagger}$ \\
\bottomrule
\end{tabular}
\\[2pt]
\footnotesize $^{\dagger}$\,Stage-2D mixed reference ($\textsc{Real}\times40 + \textsc{Cur}\times1 + \textsc{Syn}_{\text{v2 (full)}}$), trained under a different iteration budget than the other columns and included for completeness only; this reference run is distinct from the 68.12\% aggregate \emph{Real+Cur+Syn} run reported in Table~\ref{tab:main-results-comparison}.
\end{table}

\begin{table}[ht]
\centering
\small
\caption{Per-class IoU (\%) for \textbf{SegFormer} (MiT-B5). Best configuration per class in bold.}
\label{tab:perclass-segformer}
\setlength{\tabcolsep}{5pt}
\begin{tabular}{l rrrr}
\toprule
\textbf{Class} & \textbf{Real} & \textbf{+Cur} & \textbf{+Syn} & \textbf{+Cur+Syn} \\
\midrule
background         & 93.53 & 93.71          & 93.37          & \textbf{94.03} \\
elevator           & 73.07 & \textbf{76.29} & 66.76          & 72.21 \\
elevator\_button   & 76.18 & 80.75          & \textbf{84.41} & 82.60 \\
door\_button       & 72.02 & 73.83          & \textbf{77.65} & 77.60 \\
crosswalk          & 79.44 & 81.18          & \textbf{85.76} & 83.35 \\
pedestrian\_signal & 32.61 & 36.72          & \textbf{43.79} & 41.76 \\
aps\_button        & 41.45 & \textbf{42.72} & 40.95          & 42.69 \\
bus\_stop          & 77.02 & 68.77          & \textbf{80.36} & 79.98 \\
bus\_stop\_sign    & 22.16 & 19.21          & 32.20          & \textbf{35.55} \\
handrail           & 34.34 & 37.50          & \textbf{42.04} & 40.47 \\
escalator          & 69.81 & 73.13          & 73.68          & \textbf{75.29} \\
\midrule
\textbf{mean (10 FG classes)}    & 57.81 & 59.01 & 62.76 & \textbf{63.15} \\
\textbf{mean (11 classes w/ bg)} & 61.06 & 62.16 & 65.54 & \textbf{65.96} \\
\bottomrule
\end{tabular}
\end{table}

\begin{table}[ht]
\centering
\small
\caption{Per-class IoU (\%) for \textbf{DeepLabv3+} (ResNet-101). Best configuration per class in bold.}
\label{tab:perclass-dlv3}
\setlength{\tabcolsep}{5pt}
\begin{tabular}{l rrrr}
\toprule
\textbf{Class} & \textbf{Real} & \textbf{+Cur} & \textbf{+Syn} & \textbf{+Cur+Syn} \\
\midrule
background         & \textbf{93.47} & 93.46 & 93.43          & 92.78 \\
elevator           & 72.90          & 75.48 & \textbf{81.57} & 77.37 \\
elevator\_button   & \textbf{85.18} & 83.47 & 83.68          & 78.61 \\
door\_button       & 71.92          & 70.10 & \textbf{72.19} & 68.39 \\
crosswalk          & \textbf{81.12} & 78.01 & 76.44          & 76.87 \\
pedestrian\_signal & 32.28          & 31.85 & 38.73          & \textbf{40.96} \\
aps\_button        & \textbf{43.56} & 41.35 & 30.04          & 30.67 \\
bus\_stop          & \textbf{73.77} & 63.04 & 71.11          & 68.34 \\
bus\_stop\_sign    & \textbf{37.01} & 23.75 & 30.83          & 33.79 \\
handrail           & \textbf{39.79} & 27.32 & 18.56          & 24.72 \\
escalator          & \textbf{71.23} & 66.51 & 34.94          & 65.71 \\
\midrule
\textbf{mean (10 FG classes)}    & \textbf{60.88} & 56.09 & 53.81 (48.92$^\ddagger$) & 56.54 \\
\textbf{mean (11 classes w/ bg)} & \textbf{63.84} & 59.49 & 57.41 (52.63$^\ddagger$) & 59.84 \\
\bottomrule
\end{tabular}
\\[2pt]
\footnotesize $^\ddagger$\,Parenthesized values include false-positive \texttt{revolving\_door} predictions ($\text{IoU}{=}0$) in an 11/12-class denominator.
\end{table}

\begin{table}[ht]
\centering
\small
\caption{Per-class IoU (\%) for \textbf{EncNet} (ResNet-101). Best configuration per class in bold.}
\label{tab:perclass-encnet}
\setlength{\tabcolsep}{5pt}
\begin{tabular}{l rrrr}
\toprule
\textbf{Class} & \textbf{Real} & \textbf{+Cur} & \textbf{+Syn} & \textbf{+Cur+Syn} \\
\midrule
background         & 93.80          & 91.72          & \textbf{94.24} & 93.69 \\
elevator           & 77.70          & 77.18          & \textbf{79.97} & 77.19 \\
elevator\_button   & 74.34          & 72.07          & \textbf{82.71} & 81.82 \\
door\_button       & 62.57          & 60.46          & \textbf{71.89} & 69.13 \\
crosswalk          & \textbf{81.11} & 69.85          & 72.82          & 75.32 \\
pedestrian\_signal & 38.70          & \textbf{39.62} & 34.58          & 39.11 \\
aps\_button        & \textbf{43.29} & 40.02          & 34.41          & 24.30 \\
bus\_stop          & 73.20          & 0.00           & 76.41          & \textbf{77.82} \\
bus\_stop\_sign    & 25.36          & 31.34          & \textbf{38.09} & 36.83 \\
handrail           & 35.78          & 41.64          & \textbf{44.14} & 39.57 \\
escalator          & 60.31          & 56.64          & 60.78          & \textbf{63.21} \\
\midrule
\textbf{mean (10 FG classes)}    & 57.24 & 48.88 & \textbf{59.58} (54.16$^\dagger$) & 58.43 (53.12$^\dagger$) \\
\textbf{mean (11 classes w/ bg)} & 60.56 & 52.77 & \textbf{62.73} (57.50$^\dagger$) & 61.64 (56.50$^\dagger$) \\
\bottomrule
\end{tabular}
\\[2pt]
\footnotesize $^\dagger$\,Parenthesized values include false-positive \texttt{revolving\_door} predictions ($\text{IoU}{=}0$) in an 11/12-class denominator ($\Delta{=}{-}3.06$ w/ bg on \emph{+Syn} and $\Delta{=}{-}4.06$ w/ bg on \emph{+Cur+Syn} vs.\ \emph{Real}). In \emph{+Cur}, \texttt{bus\_stop} has $\text{IoU}{=}0$ and undefined precision; assigning 0 yields 62.92\% 10-class FG precision (69.91\% over 9 classes; 72.46\% w/ bg).
\end{table}

\clearpage

\section{Camera viewpoint and embodiment transfer}
\label{app:viewpoint}

Table~\ref{tab:camera_viewpoints} summarizes nominal capture settings. 

\begin{table}[h!]
\centering
\footnotesize
\setlength{\tabcolsep}{4pt}
\caption{Nominal capture settings. Go2 values describe the setup rather than per-frame calibration.}
\label{tab:camera_viewpoints}
\begin{tabular}{@{}lccc@{}}
\toprule
Embodiments & Height (m) & Pitch & HFoV \\
\midrule
Go2 & $0.45$ & $0^\circ$ & $69^\circ$ \\
Glide & $0.6$ & $-15^\circ$ / $+25^\circ$ & $85^\circ$ \\
\bottomrule
\end{tabular}
\end{table}

\section{List of generated assets}

\newpage

\section{Per-class embodiment evaluation results}
\label{app:embodiment_per_class}

This appendix complements the embodiment results in Sec.~\ref{sec:experiments} (Table~\ref{tab:embodiment}, Fig.~\ref{fig:embodiment}) with per-class IoU on the two out-of-distribution embodiment splits: the rolling guide-cane (157 images, 5 labelled classes) and the Unitree Go2 guide-dog robot (155 images, 9 labelled classes). Following the convention used elsewhere in the paper, classes with zero ground-truth instances on a split are excluded from the per-split mean. On Go2, \texttt{elevator}, \texttt{elevator\_button}, and \texttt{door\_button} each have only a single ground-truth annotation; their per-class IoU is dominated by one frame and is reported for completeness only ($\dagger$). Bold marks the best configuration per class.

\begin{table}[h!]
\centering
\small
\caption{Per-class IoU (\%) on \textbf{Glide} (157 images, 5 labelled classes).}
\label{tab:perclass-embodiment-glide}
\setlength{\tabcolsep}{4pt}
\begin{tabular}{l rrr r}
\toprule
\textbf{Class} & \textbf{SAN-ZS} & \textbf{SegFormer (Real)} & \textbf{SegFormer (Real+Syn)} & \textbf{$\Delta$} \\
\midrule
crosswalk          & 33.13 & 49.75 & \textbf{59.39} & $+$9.64 \\
pedestrian\_signal &  2.89 &  3.94 &  \textbf{4.15}  & $+$0.21 \\
aps\_button        &  0.00 &  1.44 &  \textbf{4.86}  & $+$3.42 \\
bus\_stop          &  0.36 &  0.92 &  \textbf{2.92}  & $+$2.00 \\
handrail           & 14.98 &  3.66 &  \textbf{9.51}  & $+$5.85 \\
\midrule
\textbf{mean (5 cls)} & 10.27 & 11.94 & \textbf{16.17} & $+$4.23 \\
\bottomrule
\end{tabular}
\end{table}

\begin{table}[h!]
\centering
\small
\caption{Per-class IoU (\%) on \textbf{Go2} (155 images, 9 labelled classes). $\dagger$~classes with only one ground-truth instance.}
\label{tab:perclass-embodiment-go2}
\setlength{\tabcolsep}{4pt}
\begin{tabular}{l rrr r}
\toprule
\textbf{Class} & \textbf{SAN-ZS} & \textbf{SegFormer (Real)} & \textbf{SegFormer (Real+Syn)} & \textbf{$\Delta$} \\
\midrule
elevator$^\dagger$         &  0.00 & \textbf{93.99} & 92.03 & $-$1.96 \\
elevator\_button$^\dagger$ &  0.00 &  0.00 &  0.00 &  0.00 \\
door\_button$^\dagger$     &  0.03 &  0.00 &  0.00 &  0.00 \\
crosswalk          & 43.35 & 54.13 & \textbf{55.22} & $+$1.09 \\
pedestrian\_signal & 12.07 &  7.63 & \textbf{14.99} & $+$7.36 \\
aps\_button        &  3.15 &  0.00 &  \textbf{8.50} & $+$8.50 \\
bus\_stop          &  9.32 & \textbf{25.64} & 21.08 & $-$4.56 \\
bus\_stop\_sign    &  0.00 & \textbf{11.99} &  8.59 & $-$3.40 \\
handrail           & 21.61 &  0.68 &  \textbf{7.31} & $+$6.63 \\
\midrule
\textbf{mean (9 cls)} & 9.95 & 21.56 & \textbf{23.08} & $+$1.52 \\
\bottomrule
\end{tabular}
\end{table}

\clearpage
\section{Supplementary evaluation on additional revolving-door frames}
\label{app:revolving_door}

\paragraph{Additional revolving-door evaluation setup.}
Because the original 1{,}482-image real test split contains no \texttt{revolving\_door} instances, we evaluate this category separately using 100 additional target-positive frames grouped by filename into six source videos (\texttt{V1}--\texttt{V6}). These frames were not used for task-specific training or validation. We evaluate all 28 released fine-tuned checkpoints and all five open-vocabulary zero-shot baselines without retraining or checkpoint re-selection, preserving each model's native test transforms and full class vocabulary. For text-conditioned models, we specify ``revolving door'' as the primary target prompt and separately evaluate the legacy ``turnstile'' prompt. We pool pixel-level true positives, false positives, and false negatives across all 100 frames ($119{,}018{,}518$ target pixels out of $200{,}102{,}400$ total pixels) to compute single-class IoU, precision, and recall, where pixels predicted outside the supplied target polygons count as false positives. These single-class metrics are kept strictly separate from the 10-class foreground mIoU reported on the 1{,}482-image benchmark in Table~\ref{tab:main-results-comparison}.

\paragraph{Revolving-door recognition and transfer across architectures.}
Tables~\ref{tab:revolving-door-iou}--\ref{tab:revolving-door-full} summarize performance on the 100-frame set. Among zero-shot baselines (Table~\ref{tab:revolving-door-zs}), SED-L achieves the highest pooled score at $68.94\%$ IoU ($58.47\%$ equal-video macro IoU), followed by CAT-Seg-L ($61.70\%$), SAN-ZS ($60.93\%$), SAM~3-ZS ($49.36\%$), and YOLOE-M ($20.55\%$). Among fine-tuned checkpoints (Table~\ref{tab:revolving-door-iou} and Table~\ref{tab:revolving-door-full}), SAN (\emph{Real+Cur}) achieves $68.00\%$ IoU ($57.77\%$ macro IoU). In contrast, all five closed-vocabulary architectures (DeepLabv3+, EncNet, SegFormer, Mask2Former, and SegMAN) obtain $0.00\%$ target-class IoU under \emph{Real} and \emph{Real+Cur}, which contain no positive real \texttt{revolving\_door} training examples. Adding synthetic supervision (\emph{Real+Syn} and \emph{Real+Cur+Syn}) yields non-zero predictions across all five closed-vocabulary families (with high pixel precision of $74.78\%$--$95.56\%$ where predictions occur, and reaching up to $22.42\%$ IoU on video group \texttt{V6} for SegMAN \emph{Real+Syn} and $17.90\%$ for Mask2Former \emph{Real+Syn} in Table~\ref{tab:revolving-door-videos}), but pooled IoU across all 100 frames reaches only $0.49\%$--$2.28\%$ under \emph{Real+Syn} and at most $5.66\%$ under \emph{Real+Cur+Syn} (DeepLabv3+). Thus, while synthetic augmentation improves aggregate multi-class performance on real-anchored categories in Table~\ref{tab:main-results-comparison}, it does not by itself establish reliable real-world recognition for a category lacking positive real training examples.

\begin{table}[htbp]
\centering
\small
\setlength{\tabcolsep}{6pt}
\caption{\textbf{Revolving-door IoU (\%) on the separate 100-frame positive set} across all seven fine-tuned architectures and four training-data configurations. Configuration labels follow the released checkpoint filenames. Text-conditioned models use the prompt ``revolving door''; native inference settings are retained. The constant all-foreground reference obtains $59.48\%$ IoU. Best per row in \textbf{bold}.}
\label{tab:revolving-door-iou}
\begin{tabular}{lrrrr}
\toprule
\textbf{Model} & \textbf{Real} & \textbf{Real+Cur} & \textbf{Real+Syn} & \textbf{Real+Cur+Syn} \\
\midrule
SegMAN~\citep{segman}            & 0.00          & 0.00           & \textbf{1.10} & 0.67 \\
Mask2Former~\citep{mask2former}  & 0.00          & 0.00           & \textbf{1.38} & 0.74 \\
SAM~3~\citep{sam3}               & \textbf{27.24}& 26.15          & 15.14         & 12.99 \\
SAN~\citep{san}                  & 64.92         & \textbf{68.00} & 4.98          & 27.93 \\
SegFormer~\citep{segformer}      & 0.00          & 0.00           & \textbf{2.28} & 0.84 \\
EncNet~\citep{encnet}            & 0.00          & 0.00           & 0.75          & \textbf{4.88} \\
DeepLabv3+~\citep{deeplabv3plus} & 0.00          & 0.00           & 0.49          & \textbf{5.66} \\
\bottomrule
\end{tabular}
\end{table}

\begin{table}[htbp]
\centering
\small
\setlength{\tabcolsep}{6pt}
\caption{\textbf{Zero-shot revolving-door results (\%)} on the same 100 positive frames using the target prompt ``revolving door.'' CAT-Seg and SED retain their published background vocabulary; YOLOE and SAM~3 use confidence thresholds $0.05$ and $0.25$, respectively (SAM~3 uses FP32 image-processor inference here; the native BF16 batched evaluator yields $49.42\%$ IoU). The final row reports a constant all-foreground reference rather than a learned model.}
\label{tab:revolving-door-zs}
\begin{tabular}{lrrr}
\toprule
\textbf{Model} & \textbf{IoU} & \textbf{Prec} & \textbf{Rec} \\
\midrule
SAN-ZS~\citep{san}         & 60.93          & 62.00          & \textbf{97.25} \\
CAT-Seg-L~\citep{catseg}   & 61.70          & 83.63          & 70.18 \\
SED-L~\citep{sed}          & \textbf{68.94} & 75.91          & 88.24 \\
YOLOE-M~\citep{yoloe}      & 20.55          & 85.71          & 21.28 \\
SAM~3-ZS~\citep{sam3}      & 49.36          & \textbf{96.59} & 50.23 \\
\midrule
All pixels foreground (ref.) & 59.48        & 59.48          & 100.00 \\
\bottomrule
\end{tabular}
\end{table}

\paragraph{Indirect open-vocabulary transfer and prompt sensitivity.}
Notably, SAN improves from $60.93\%$ zero-shot IoU to $64.92\%$ under \emph{Real} and $68.00\%$ under \emph{Real+Cur}, despite the absence of positive real \texttt{revolving\_door} training instances in those splits. Because all three SAN runs share identical prompt vocabularies, preprocessing, and inference thresholds, this gain reflects reduced over-segmentation after adapting shared mask and background representations on the other 10 \textsc{NavAble} classes: relative to SAN-ZS, SAN (\emph{Real+Cur}) reduces false-positive pixels from $70.95$~million to $38.94$~million (lifting precision from $62.00\%$ to $73.39\%$) while recall shifts from $97.25\%$ to $90.25\%$. Conversely, SAM~3 declines from $49.36\%$ (zero-shot) to $27.24\%$ (\emph{Real}); however, because published zero-shot and fine-tuned SAM~3 recipes use different confidence thresholds ($0.25$ vs.\ $0.50$) and overlap-resolution rules, this difference represents a system-level comparison rather than an isolated causal test of weight updates. Furthermore, Table~\ref{tab:revolving-door-full}b shows that open-vocabulary models are highly sensitive to class wording: querying the exact same checkpoints with the legacy token ``turnstile'' instead of ``revolving door'' reduces IoU to $0.00\%$--$5.52\%$ across all evaluated models.

\begin{table}[htbp]
\centering
\footnotesize
\setlength{\tabcolsep}{5pt}
\caption{\textbf{Full revolving-door results (\%,left) and target-prompt sensitivity (\%, right)} on the 100-frame set. (Left) Pixel-level precision, recall, and IoU for all 28 released fine-tuned checkpoints; a dash (\texttt{--}) indicates no target pixels were predicted. (Right) Pooled IoU when querying identical weights with ``revolving door'' versus the legacy token ``turnstile.''}
\label{tab:revolving-door-full}
\begin{adjustbox}{max width=\textwidth}
\begin{tabular}{llrrr @{\hspace{18pt}} llrr}
\toprule
\multicolumn{5}{c}{\textbf{(a) All 28 Fine-Tuned Checkpoints (``revolving door'')}} & \multicolumn{4}{c}{\textbf{(b) Target-Prompt Sensitivity (IoU, \%)}} \\
\cmidrule(r){1-5} \cmidrule(l){6-9}
\textbf{Model} & \textbf{Config} & \textbf{Prec} & \textbf{Rec} & \textbf{IoU} & \textbf{Model} & \textbf{Configuration} & \textbf{Revolving door} & \textbf{Turnstile} \\
\midrule
SegMAN      & Real         & --    &  0.00 &  0.00 & SAM~3     & Real         & 27.24 & 0.00 \\
SegMAN      & Real+Cur     & --    &  0.00 &  0.00 & SAM~3     & Real+Cur     & 26.15 & 0.00 \\
SegMAN      & Real+Syn     & 94.96 &  1.10 &  1.10 & SAM~3     & Real+Syn     & 15.14 & 0.00 \\
SegMAN      & Real+Cur+Syn & 94.29 &  0.67 &  0.67 & SAM~3     & Real+Cur+Syn & 12.99 & 0.35 \\
Mask2Former & Real         & --    &  0.00 &  0.00 & SAN       & Real         & 64.92 & 0.00 \\
Mask2Former & Real+Cur     & --    &  0.00 &  0.00 & SAN       & Real+Cur     & 68.00 & 0.00 \\
Mask2Former & Real+Syn     & 93.07 &  1.38 &  1.38 & SAN       & Real+Syn     &  4.98 & 3.30 \\
Mask2Former & Real+Cur+Syn & 95.56 &  0.74 &  0.74 & SAN       & Real+Cur+Syn & 27.93 & 1.54 \\
SAM~3       & Real         & 99.78 & 27.26 & 27.24 & SAN-ZS    & Zero-shot    & 60.93 & 2.46 \\
SAM~3       & Real+Cur     & 99.53 & 26.18 & 26.15 & CAT-Seg-L & Zero-shot    & 61.70 & 0.00 \\
SAM~3       & Real+Syn     & 96.99 & 15.22 & 15.14 & SED-L     & Zero-shot    & 68.94 & 5.52 \\
SAM~3       & Real+Cur+Syn & 99.84 & 13.00 & 12.99 & YOLOE-M   & Zero-shot    & 20.55 & 0.17 \\
SAN         & Real         & 74.48 & 83.49 & 64.92 & SAM~3-ZS  & Zero-shot    & 49.36 & 0.00 \\
SAN         & Real+Cur     & 73.39 & 90.25 & 68.00 &           &              &       &      \\
SAN         & Real+Syn     & 77.95 &  5.06 &  4.98 &           &              &       &      \\
SAN         & Real+Cur+Syn & 80.19 & 30.00 & 27.93 &           &              &       &      \\
SegFormer   & Real         & --    &  0.00 &  0.00 &           &              &       &      \\
SegFormer   & Real+Cur     & --    &  0.00 &  0.00 &           &              &       &      \\
SegFormer   & Real+Syn     & 84.99 &  2.29 &  2.28 &           &              &       &      \\
SegFormer   & Real+Cur+Syn & 79.15 &  0.84 &  0.84 &           &              &       &      \\
EncNet      & Real         & --    &  0.00 &  0.00 &           &              &       &      \\
EncNet      & Real+Cur     & --    &  0.00 &  0.00 &           &              &       &      \\
EncNet      & Real+Syn     & 90.60 &  0.75 &  0.75 &           &              &       &      \\
EncNet      & Real+Cur+Syn & 85.87 &  4.92 &  4.88 &           &              &       &      \\
DeepLabv3+  & Real         & --    &  0.00 &  0.00 &           &              &       &      \\
DeepLabv3+  & Real+Cur     & --    &  0.00 &  0.00 &           &              &       &      \\
DeepLabv3+  & Real+Syn     & 94.19 &  0.49 &  0.49 &           &              &       &      \\
DeepLabv3+  & Real+Cur+Syn & 74.78 &  5.77 &  5.66 &           &              &       &      \\
\bottomrule
\end{tabular}
\end{adjustbox}
\end{table}

\begin{figure}[htbp]
\centering
\includegraphics[width=0.93\linewidth]{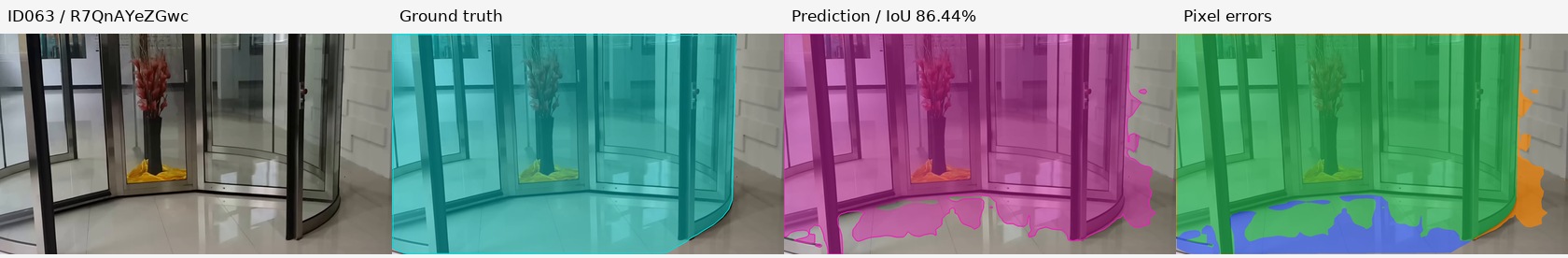}\\[3pt]
\includegraphics[width=0.99\linewidth]{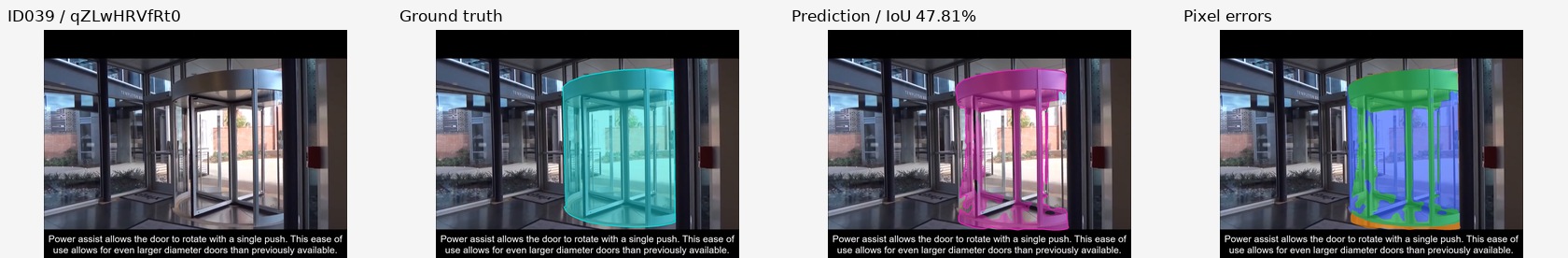}
\caption{\textbf{Representative revolving-door predictions.} (Top) SAN (\emph{Real+Cur}) on frame \texttt{ID063} (\texttt{V4}, $86.44\%$ IoU) captures the full door enclosure. (Bottom) SegMAN (\emph{Real+Syn}) on frame \texttt{ID039} (\texttt{V6}, $47.81\%$ IoU) segments the metallic door frame/leaves while omitting transparent glass regions included in the ground-truth polygon (cyan: ground truth; magenta: prediction; green/orange/blue: TP/FP/FN).}
\label{fig:revolving-door-qual}
\end{figure}

\paragraph{Scope, source correlation, and annotation limitations.}
Several factors constrain the interpretation of this supplementary set. First, all 100 frames are target-positive and come from six correlated video sequences (\texttt{V1}: 16 frames, $32.80\%$ target pixels; \texttt{V2}: 10 frames, $34.24\%$; \texttt{V3}: 6 frames, $96.37\%$; \texttt{V4}: 34 frames, $90.11\%$; \texttt{V5}: 20 frames, $47.76\%$; \texttt{V6}: 14 frames, $26.23\%$). Because target masks cover $59.48\%$ of all pixels across the set ($54.59\%$ equal-video average), predicting every pixel as foreground yields $59.48\%$ pooled IoU without visual recognition, and positive-only sampling does not measure false alarms on target-absent scenes. Table~\ref{tab:revolving-door-videos} therefore reports per-video IoU and unweighted 6-video macro averages alongside pooled scores. Second, qualitative inspection indicates a boundary-definition gap between sources (illustrated in Figure~\ref{fig:revolving-door-qual}): the supplied ground-truth polygons label the full projected door enclosure (including transparent glass panels, interior floor regions, and background visible through the glass), whereas synthetic-trained closed-vocabulary models primarily fire on rigid metallic door frames and leaves. Finally, unlike the main collected dataset (which underwent our 100-image independent reannotation audit achieving $97.2\%$ pairwise mask mIoU in Appendix~\ref{app:annotation_filtering}), no independent inter-annotator agreement audit was conducted on these 100 supplementary frames.

\begin{table}[htbp]
\centering
\scriptsize
\setlength{\tabcolsep}{4.5pt}
\renewcommand{\arraystretch}{1.05}
\caption{\textbf{Revolving-door IoU (\%) by filename-derived source-video group (\texttt{V1}--\texttt{V6}).} \textbf{Macro} reports the unweighted mean across the six video groups as a sensitivity check against group size and foreground occupancy.}
\label{tab:revolving-door-videos}
\begin{adjustbox}{max width=\textwidth}
\begin{tabular}{lrrrrrrr}
\toprule
\textbf{Model / Configuration} & \textbf{V1} & \textbf{V2} & \textbf{V3} & \textbf{V4} & \textbf{V5} & \textbf{V6} & \textbf{Macro} \\
\midrule
SegMAN / Real              &  0.00 &  0.00 &  0.00 &  0.00 &  0.00 &  0.00 &  0.00 \\
SegMAN / Real+Cur          &  0.00 &  0.00 &  0.00 &  0.00 &  0.00 &  0.00 &  0.00 \\
SegMAN / Real+Syn          &  0.12 &  0.00 &  0.00 &  0.00 &  0.00 & 22.42 &  3.76 \\
SegMAN / Real+Cur+Syn      &  0.00 &  0.00 &  0.00 &  0.04 &  0.00 & 13.52 &  2.26 \\
Mask2Former / Real         &  0.00 &  0.00 &  0.00 &  0.00 &  0.00 &  0.00 &  0.00 \\
Mask2Former / Real+Cur     &  0.00 &  0.00 &  0.00 &  0.00 &  0.00 &  0.00 &  0.00 \\
Mask2Former / Real+Syn     &  0.66 &  0.00 &  0.00 &  0.67 &  0.59 & 17.90 &  3.30 \\
Mask2Former / Real+Cur+Syn &  4.08 &  0.00 &  0.00 &  0.39 &  0.14 &  2.78 &  1.23 \\
SAM~3 / Real               & 83.06 & 47.71 &  0.00 & 19.19 & 29.62 & 33.64 & 35.54 \\
SAM~3 / Real+Cur           & 85.98 & 46.64 &  0.00 & 13.82 & 41.51 & 24.46 & 35.40 \\
SAM~3 / Real+Syn           & 48.75 & 34.20 &  0.00 &  4.01 & 34.97 & 12.34 & 22.38 \\
SAM~3 / Real+Cur+Syn       & 60.03 & 25.36 &  0.00 &  0.00 & 27.58 & 29.08 & 23.68 \\
SAN / Real                 & 43.43 & 50.17 & 65.36 & 80.28 & 60.50 & 47.81 & 57.92 \\
SAN / Real+Cur             & 41.44 & 53.02 & 49.02 & 92.27 & 57.71 & 53.19 & 57.77 \\
SAN / Real+Syn             & 32.78 &  4.51 &  0.00 &  0.04 &  5.82 & 10.16 &  8.89 \\
SAN / Real+Cur+Syn         & 50.37 & 20.67 &  0.00 & 28.09 & 36.00 & 20.39 & 25.92 \\
SegFormer / Real           &  0.00 &  0.00 &  0.00 &  0.00 &  0.00 &  0.00 &  0.00 \\
SegFormer / Real+Cur       &  0.00 &  0.00 &  0.00 &  0.00 &  0.00 &  0.00 &  0.00 \\
SegFormer / Real+Syn       &  1.84 &  0.00 &  1.82 &  2.48 &  0.00 & 11.81 &  2.99 \\
SegFormer / Real+Cur+Syn   &  0.00 &  0.00 &  0.00 &  0.01 &  0.00 & 16.74 &  2.79 \\
EncNet / Real              &  0.00 &  0.00 &  0.00 &  0.00 &  0.00 &  0.00 &  0.00 \\
EncNet / Real+Cur          &  0.00 &  0.00 &  0.00 &  0.00 &  0.00 &  0.00 &  0.00 \\
EncNet / Real+Syn          &  1.14 &  0.00 &  0.03 &  1.10 &  0.00 &  1.11 &  0.56 \\
EncNet / Real+Cur+Syn      &  2.12 &  0.07 &  0.74 &  8.47 &  0.00 &  1.18 &  2.10 \\
DeepLabv3+ / Real          &  0.00 &  0.00 &  0.00 &  0.00 &  0.00 &  0.00 &  0.00 \\
DeepLabv3+ / Real+Cur      &  0.00 &  0.00 &  0.00 &  0.00 &  0.00 &  0.00 &  0.00 \\
DeepLabv3+ / Real+Syn      &  1.73 &  0.01 &  0.00 &  0.09 &  0.00 &  5.83 &  1.28 \\
DeepLabv3+ / Real+Cur+Syn  &  1.46 &  0.00 &  0.00 &  9.58 &  0.00 &  7.08 &  3.02 \\
\midrule
SAN-ZS / Zero-shot         & 34.18 & 39.38 & 74.54 & 89.72 & 49.16 & 33.08 & 53.34 \\
CAT-Seg-L / Zero-shot      & 65.87 & 51.38 & 13.59 & 70.08 & 67.86 & 54.66 & 53.90 \\
SED-L / Zero-shot          & 56.10 & 54.35 & 53.59 & 90.13 & 61.60 & 35.06 & 58.47 \\
YOLOE-M / Zero-shot        & 48.37 & 21.93 & 11.03 &  6.79 & 46.07 & 36.65 & 28.48 \\
SAM~3-ZS / Zero-shot       & 87.78 & 48.47 &  0.00 & 47.32 & 57.99 & 71.04 & 52.10 \\
\midrule
All pixels foreground (ref.) & 32.80 & 34.24 & 96.37 & 90.11 & 47.76 & 26.23 & 54.59 \\
\bottomrule
\end{tabular}
\end{adjustbox}
\\[2pt]
{\footnotesize Video groups: \texttt{V1}: \texttt{5UmmMBssTRA} (16 frames); \texttt{V2}: \texttt{DVwbq0AacR4} (10 frames); \texttt{V3}: \texttt{F3-3BXJhouw} (6 frames); \texttt{V4}: \texttt{R7QnAYeZGwc} (34 frames); \texttt{V5}: \texttt{UOtBRjNm7nY} (20 frames); \texttt{V6}: \texttt{qZLwHRVfRt0} (14 frames).}
\end{table}

\clearpage
\section{Candidate Public Repositories Surveyed for Data Curation}
\label{app:curation_sources}

Tables~\ref{tab:curation-handrail}--\ref{tab:curation-bus-stop-sign} list the candidate public repositories surveyed across open-source hosting platforms (Roboflow Universe, Kaggle, Mapillary, Mendeley Data, and university repositories) during the initial source-level screening phase of \textsc{NavAble} curation (Appendix~\ref{app:annotation_filtering}). Importantly, this list represents the raw candidate pool prior to applying our seven source-level exclusion criteria: the majority of surveyed projects were excluded at the source level because they provided only bounding-box detection (\texttt{detec}) or image-level classification (\texttt{classify}) labels without pixel-level segmentation masks, contained unannotated images (\texttt{image\_only}), duplicated other merged repositories, or focused on non-target vehicle traffic signage. Only repositories providing compatible polygon/mask annotations for our target taxonomy were retained to form the $45{,}060$-image downloaded segmentation pool, which was subsequently filtered via exhaustive manual mask-overlay inspection to yield the $35{,}825$-image initial filtered pool (Table~\ref{tab:annotation_filtering_statistics}; $36{,}466$ total unique images in the final released \textsc{NavAble}\(_{\text{cur}}\) split in Table~\ref{tab:per_class_counts}).

\paragraph{Data availability and redistribution.}
For third-party imagery retained in \textsc{NavAble} and subject
to redistribution restrictions, we release only our annotations.
Readers should obtain the corresponding images directly from
the original sources linked in the tables below.

\subsection*{Handrail}
\vspace{-2mm}
\begin{center}
\scriptsize
\setlength{\tabcolsep}{4pt}
\renewcommand{\arraystretch}{1.05}
\captionof{table}{Candidate public repositories surveyed for \texttt{handrail}.}
\label{tab:curation-handrail}
\begin{tabularx}{\textwidth}{r p{3.6cm} l X}
\toprule
\textbf{\#} & \textbf{Project Name} & \textbf{Type} & \textbf{URL} \\
\midrule
1 & handrail & detec & \url{https://universe.roboflow.com/da-nsoap/handrail/dataset/12} \\
2 & Stairs -- Handrail Detector & detec & \url{https://universe.roboflow.com/itlabroom/stairs-handrail-detector/dataset/2} \\
3 & hand-desk-tab & seg & \url{https://universe.roboflow.com/visionaotwo/hand-desk-tab/browse} \\
4 & Handrail detection v2 & detec & \url{https://universe.roboflow.com/vnexia-internship/handrail-detection-v2-myw0q/dataset/11} \\
5 & handrail & seg & \url{https://universe.roboflow.com/visionaotwo/handrail-9nbmww} \\
6 & PathosV2 & detec & \url{https://universe.roboflow.com/pathos/pathosv2/dataset/1} \\
\bottomrule
\end{tabularx}
\end{center}
\vspace{-2mm}
\begin{itemize}
    \item \textbf{Source platform:} All surveyed repositories are hosted on Roboflow Universe.
    \item \textbf{Curation notes:} Rows~1, 2, 4, and~6 provide bounding-box detection annotations (with Rows~2 and~4 co-labeling stairs and Row~6 containing a partial subset of handrail images); Rows~3 and~5 are segmentation projects noted as \texttt{image\_only} where usable masks were screened during curation.
\end{itemize}

\subsection*{Escalator}
\vspace{-2mm}
\begin{center}
\scriptsize
\setlength{\tabcolsep}{4pt}
\renewcommand{\arraystretch}{1.05}
\captionof{table}{Candidate public repositories surveyed for \texttt{escalator}.}
\label{tab:curation-escalator}
\begin{tabularx}{\textwidth}{r p{3.6cm} l X}
\toprule
\textbf{\#} & \textbf{Project Name} & \textbf{Type} & \textbf{URL} \\
\midrule
1 & Escalator Data & seg & \url{https://universe.roboflow.com/escalator-dataset/escalator-data/dataset/1} \\
2 & Escalator Falling & detec & \url{https://universe.roboflow.com/sumitesh-naithani-3ijrc/escalator-falling/dataset/11} \\
3 & Escalator & detec & \url{https://universe.roboflow.com/mallescalatorsystem/escalator-vsvly/dataset/1} \\
4 & PF & detec & \url{https://universe.roboflow.com/pf-mikpj/pf-83mz4/dataset/1} \\
5 & SubwayObjectDetection & detec & \url{https://universe.roboflow.com/seoultech-8im0w/subwayobjectdetection-o5q0n/dataset/2} \\
6 & merge & detec & \url{https://universe.roboflow.com/deep-ksl1a/merge-hypmm/browse?queryText=class\%3Aescalator&pageSize=50&startingIndex=0&browseQuery=true} \\
7 & final\_DATA & detec & \url{https://universe.roboflow.com/deep-ksl1a/final_data-ifwvq/dataset/2} \\
8 & escalator & detec & \url{https://universe.roboflow.com/mess-w7qn0/escalator-cez3t/browse?queryText=&pageSize=50&startingIndex=0&browseQuery=true} \\
\bottomrule
\end{tabularx}
\end{center}
\vspace{-2mm}
\begin{itemize}
    \item \textbf{Source platform:} All surveyed repositories are hosted on Roboflow Universe.
    \item \textbf{Curation notes:} Row~1 provides segmentation masks; Rows~2--8 are detection-only or \texttt{image\_only} (Row~8) repositories, with Rows~5 and~7 also containing \texttt{elevator} and \texttt{toilet} classes.
\end{itemize}

\subsection*{Revolving Door / Turnstile}
\vspace{-2mm}
\begin{center}
\scriptsize
\setlength{\tabcolsep}{4pt}
\renewcommand{\arraystretch}{1.05}
\captionof{table}{Candidate public repositories surveyed for \texttt{revolving\_door}.}
\label{tab:curation-revolving-door}
\begin{tabularx}{\textwidth}{r p{3.6cm} l X}
\toprule
\textbf{\#} & \textbf{Project Name} & \textbf{Type} & \textbf{URL} \\
\midrule
1 & YOLOTurnstile & detec & \url{https://universe.roboflow.com/uatrainrecognition/yoloturnstile/dataset/4} \\
2 & ingong & detec & \url{https://universe.roboflow.com/spspspspsp/ingong-aoabg/dataset/3} \\
3 & fr & detec & \url{https://universe.roboflow.com/spspspspsp/fr-cnruk/dataset/2} \\
4 & Roboflow Object Identification & seg & \url{https://universe.roboflow.com/project/roboflow-object-identification-gix9j/dataset/11/images?split=train} \\
\bottomrule
\end{tabularx}
\end{center}
\vspace{-2mm}
\begin{itemize}
    \item \textbf{Source platform:} All surveyed repositories are hosted on Roboflow Universe.
    \item \textbf{Curation notes:} Rows~1--3 provide bounding-box detection only (with Rows~2 and~3 also labeling \texttt{person} and \texttt{hand}); Row~4 includes both \texttt{revolving\_door} and generic \texttt{door} labels, and none yielded retained real-world segmentation instances after quality screening.
\end{itemize}

\clearpage
\subsection*{Crosswalk}
\vspace{-2mm}
\begin{center}
\scriptsize
\setlength{\tabcolsep}{3.5pt}
\renewcommand{\arraystretch}{1.02}
\captionof{table}{Candidate public repositories surveyed for \texttt{crosswalk}.}
\label{tab:curation-crosswalk}
\begin{tabularx}{\textwidth}{r p{3.6cm} l X}
\toprule
\textbf{\#} & \textbf{Project Name} & \textbf{Type} & \textbf{URL} \\
\midrule
1 & crosswalk-dataset & classify & \url{https://www.kaggle.com/datasets/davidsilvam/crosswalkdataset} \\
2 & crosswalk-seg & seg & \url{https://universe.roboflow.com/tcc-xn3st/crosswalk-seg/browse?queryText=&pageSize=50&startingIndex=0&browseQuery=true} \\
3 & crosswalk & detec & \url{https://universe.roboflow.com/tcc-xn3st/crosswalk-djojj} \\
4 & crosswalk-TA & seg & \url{https://universe.roboflow.com/image-process-1mqey/crosswalk-ta/dataset/8} \\
5 & JayWatch7 & seg & \url{https://universe.roboflow.com/jaywatch-3cown/jaywatch7-yg7zd} \\
6 & crosswalk & detec & \url{https://universe.roboflow.com/wqwdas/crosswalk-1elwe/dataset/2} \\
7 & zebra cross TA & detec & \url{https://universe.roboflow.com/image-process-1mqey/zebra-cross-ta/dataset/2} \\
8 & crosswalk and motorcycle & detec & \url{https://universe.roboflow.com/barton-hill-blind-walking-stick/crosswalk-and-motorcycle/browse?queryText=class\%3Acrosswalk&pageSize=50&startingIndex=0&browseQuery=true} \\
9 & Grand Army & detec & \url{https://universe.roboflow.com/dot-hr21x/grand-army/dataset/11/images?split=train} \\
10 & Crossy road & detec & \url{https://universe.roboflow.com/sixthsense/crossy-road-c8gfs} \\
11 & tak{\i}m & detec & \url{https://universe.roboflow.com/etu-tx9q2/takim/dataset/2} \\
12 & crosswalk\_2 & detec & \url{https://universe.roboflow.com/testestest/crosswalk_2-yuh40/dataset/5} \\
13 & safeT\_Crosswalk & detec & \url{https://universe.roboflow.com/mia-lkjrc/safet_crosswalk/dataset/11} \\
14 & test & detec & \url{https://universe.roboflow.com/etu-tx9q2/test-lepfa/dataset/3} \\
15 & AuralVision & detec & \url{https://universe.roboflow.com/merlimar-zt8am/auralvision/dataset/3} \\
16 & crosswalk-final & detec & \url{https://universe.roboflow.com/tcc-xn3st/crosswalk-final/dataset/1} \\
17 & Road-users-disabilities & detec & \url{https://universe.roboflow.com/project/road-users-disabilities/dataset/5} \\
18 & crossWalkDetect & detec & \url{https://universe.roboflow.com/onetwo/crosswalkdetect/dataset/4} \\
19 & crossing\_street & seg & \url{https://universe.roboflow.com/ming-home/crossing_street/dataset/11} \\
20 & pedestrian light, crosswalk & detec & \url{https://universe.roboflow.com/chanyoung/pedestrian-light-crosswalk/dataset/2} \\
21 & walk-cross & detec & \url{https://universe.roboflow.com/test-jyqch/walk-cross/dataset/1} \\
22 & WTF & detec & \url{https://universe.roboflow.com/ajsh-yai1y/wtf-dwvgm/dataset/2} \\
23 & Crosswalk identifier & detec & \url{https://universe.roboflow.com/envs333-proj/crosswalk-identifier/dataset/10} \\
24 & huh & detec & \url{https://universe.roboflow.com/traffic-signs-vm8gh/huh-scajc/dataset/1} \\
25 & SafeCross & seg & \url{https://universe.roboflow.com/imt-mines-als-by61v/safecross-5uowm/dataset/2} \\
26 & crosswalk & seg & \url{https://universe.roboflow.com/lane-detection-on23z/crosswalk-ukc2k/dataset/1} \\
27 & safecross-filtered & detec & \url{https://universe.roboflow.com/saadi-space/safecross-filtered-qiokr/dataset/2} \\
28 & CrosswalkAndWaitingZone & seg & \url{https://universe.roboflow.com/teste1-j7efq/crosswalkandwaitingzone} \\
29 & lane marking-v1 & seg & \url{https://universe.roboflow.com/inc-hymh5/lane-marking-v1/dataset/5} \\
30 & combine & detec & \url{https://universe.roboflow.com/saadfl/combine-jcyrk/browse} \\
31 & street-obj-extra & detec & \url{https://universe.roboflow.com/s-workspace-ojshb/street-obj-extra} \\
32 & crosswalk & seg & \url{https://universe.roboflow.com/crosswalk-fx95c/crosswalk-bhgbx/dataset/1} \\
33 & crosswalk deteccting & detec & \url{https://universe.roboflow.com/project-sgd9c/crosswalk-deteccting} \\
34 & detec & detec & \url{https://universe.roboflow.com/ajjs-kpvfj/detect-sibjy/dataset/1} \\
35 & cross\_walk & detec & \url{https://universe.roboflow.com/new-workspace-jbi8t/cross_walk/dataset/7} \\
36 & CrosSight & detec & \url{https://universe.roboflow.com/crossight/crossight/dataset/20} \\
37 & FinalProject1 & detec & \url{https://universe.roboflow.com/school-sacst/finalproject1} \\
38 & zebra crossing & detec & \url{https://universe.roboflow.com/zebra-crossing-1qzhk/zebra-crossing-awdrp/dataset/1} \\
39 & yolov8-crosswalk-detection & detec & \url{https://universe.roboflow.com/tcc-xn3st/yolov8-crosswalk-detection/browse?queryText=&pageSize=50&startingIndex=0&browseQuery=true} \\
40 & CrossWalk & seg & \url{https://universe.roboflow.com/khiem-mye7k/crosswalk-mga2z/dataset/1} \\
\bottomrule
\end{tabularx}
\end{center}
\vspace{-2mm}
\begin{itemize}
    \item \textbf{Source platforms:} Row~1 is hosted on Kaggle; Rows~2--40 are hosted on Roboflow Universe.
    \item \textbf{Curation notes:} Row~1 contains classification labels only (no bounding boxes or masks); Rows~2, 4, 5, 19, 25, 26, 28, 29, 32, and~40 provide segmentation annotations (with Row~29 also including \texttt{stop-line}), whereas the remaining projects are bounding-box detection sets or \texttt{image\_only} (Row~37), several of which co-label vehicles, motorcycles, pedestrians, or pedestrian signals (Rows~7, 8, 15, 17, 18, 20, 30, 36).
\end{itemize}

\clearpage
\subsection*{Traffic Signal}
\vspace{-2mm}
\begin{center}
\scriptsize
\setlength{\tabcolsep}{3.5pt}
\renewcommand{\arraystretch}{1.02}
\captionof{table}{Candidate public repositories surveyed for general traffic signals and signage.}
\label{tab:curation-traffic-signal}
\begin{tabularx}{\textwidth}{r p{3.6cm} l X}
\toprule
\textbf{\#} & \textbf{Project Name} & \textbf{Type} & \textbf{URL} \\
\midrule
1 & Mapillary Traffic Sign Dataset & detec & \url{https://www.mapillary.com/dataset/trafficsign} \\
2 & Traffic Signs Detection & detec & \url{https://www.kaggle.com/datasets/pkdarabi/cardetection} \\
3 & Object\_Detection\_Final\_Dataset & detec & \url{https://universe.roboflow.com/traffic-sign-3qmuf/object_detection_final_dataset-eqomr} \\
4 & Segmentation\_Final\_Dataset & seg & \url{https://universe.roboflow.com/traffic-sign-3qmuf/segmentation_final_dataset} \\
5 & Trafic Signal & detec & \url{https://universe.roboflow.com/object-h7949/trafic-signal} \\
6 & Trafic signal light & detec & \url{https://universe.roboflow.com/mohammed-owais-dr4nb/trafic-signal-light} \\
7 & Trafic signal light & detec & \url{https://universe.roboflow.com/trafficsign-eipbs/traffic-sign-qrgpa} \\
8 & traffic-signs-segmentation1 & seg & \url{https://universe.roboflow.com/traffic-signs-segmentation/traffic-signs-segmentation1/dataset/7} \\
9 & traffic & detec & \url{https://universe.roboflow.com/project/traffic-u0cfy/dataset/1/images?split=train} \\
10 & Traffic Signs & detec & \url{https://www.cvl.isy.liu.se/research/datasets/traffic-signs-dataset/} \\
11 & Mapillary Vistas Dataset & detec & \url{https://www.mapillary.com/dataset/vistas} \\
12 & traffic sign detection & detec & \url{https://universe.roboflow.com/jy-ij8pd/traffic-sign-detection-yknyn} \\
13 & Team Black All Traffic Signs & seg & \url{https://universe.roboflow.com/jy-ij8pd/team-black-all-traffic-signs/dataset/8} \\
14 & street view traffic signs & detec & \url{https://universe.roboflow.com/haraldtan/street-view-traffic-signs/dataset/4/images?split=train} \\
15 & speed limit signs & detec & \url{https://universe.roboflow.com/haraldtan/speed-limit-signs-xlyo3/dataset/1} \\
16 & Junction Detection & seg & \url{https://universe.roboflow.com/muhammad-asyraf-mohd-zain-qwgwv/junction-detection-vrysn/dataset/3} \\
17 & emtd & detec & \url{https://universe.roboflow.com/traffic-sign-3mv8c/emtd/browse?queryText=&pageSize=50&startingIndex=0&browseQuery=true} \\
18 & Traffic SignRGB & detec & \url{https://universe.roboflow.com/project/traffic-signrgb/dataset/2/images?split=train} \\
19 & For report writing & detec & \url{https://universe.roboflow.com/final-year-project-bk8o4/for-report-writing} \\
20 & traffic signs\_04 & detec & \url{https://universe.roboflow.com/project-ypnwz/traffic-signs_04} \\
21 & traffic signs merged & detec & \url{https://universe.roboflow.com/project-ypnwz/traffic-signs-merged/browse?queryText=&pageSize=50&startingIndex=0&browseQuery=true} \\
22 & traffic signs merged 2 & detec & \url{https://universe.roboflow.com/project-ypnwz/traffic-signs-merged-2/dataset/3} \\
23 & traffic signs final & detec & \url{https://universe.roboflow.com/project-ypnwz/traffic-signs-final/browse?queryText=&pageSize=200&startingIndex=0&browseQuery=true} \\
24 & traffic signs-02 & detec & \url{https://universe.roboflow.com/project-ypnwz/traffic-signs-02/dataset/2} \\
25 & traffic signs & detec & \url{https://universe.roboflow.com/project-ypnwz/traffic-signs-92sv3/browse?queryText=&pageSize=50&startingIndex=0&browseQuery=true} \\
26 & Keep Left Signs & seg & \url{https://universe.roboflow.com/jy-ij8pd/keep-left-signs/dataset/4/images?split=train} \\
27 & Traffic Light Detection & detec & \url{https://universe.roboflow.com/fyp-project-1n6xu/traffic-light-detection-0i55w} \\
28 & Traffic Lights 3 & detec & \url{https://universe.roboflow.com/zhengrong616-hotmail-com/traffic-lights-3/dataset/2} \\
29 & yolov8n\_preliminaryTest & detec & \url{https://universe.roboflow.com/fyp-wnch0/yolov8n_preliminarytest/dataset/4/images?split=train} \\
30 & traffic\_lights\_red\_green\_man & detec & \url{https://universe.roboflow.com/fyp-axpvy/traffic_lights_red_green_man/dataset/2} \\
31 & traffic\_lights & detec & \url{https://universe.roboflow.com/fyp-axpvy/traffic_lights-ndzdd/dataset/6} \\
32 & PhaseB\_Augmentation & detec & \url{https://universe.roboflow.com/dcil-b30bp/phaseb_augmentation/dataset/3} \\
33 & Traffic-Signs-Dataset & detec & \url{https://www.kaggle.com/datasets/tuanai/traffic-signs-dataset} \\
\bottomrule
\end{tabularx}
\end{center}
\vspace{-2mm}
\begin{itemize}
    \item \textbf{Source platforms:} Rows~1 and~11 are from Mapillary; Rows~2 and~33 are from Kaggle; Row~10 is from Link\"oping University CVL; all remaining repositories are hosted on Roboflow Universe.
    \item \textbf{Curation notes:} Most repositories in this category focus on general vehicular traffic signs (e.g., stop signs in Rows~12 and~18, speed-limit or keep-left signs in Rows~15 and~26, road markings in Row~4, unannotated/partially labeled sets in Rows~6 and~10, or non-English class labels in Row~8) or bounding-box detection, and were excluded when standardizing onto our pedestrian-focused taxonomy.
\end{itemize}

\clearpage
\subsection*{Pedestrian Signal}
\vspace{-2mm}
\begin{center}
\scriptsize
\setlength{\tabcolsep}{4pt}
\renewcommand{\arraystretch}{1.04}
\captionof{table}{Candidate public repositories surveyed for \texttt{pedestrian\_signal}.}
\label{tab:curation-pedestrian-signal}
\begin{tabularx}{\textwidth}{r p{3.6cm} l X}
\toprule
\textbf{\#} & \textbf{Project Name} & \textbf{Type} & \textbf{URL} \\
\midrule
1 & pedestrian traffic light & detec & \url{https://universe.roboflow.com/cible/pedestrian-traffic-light-3p4dd/dataset/1} \\
2 & pedestrian-traffic-light & detec & \url{https://universe.roboflow.com/project-wdkej/pedestrian-traffic-light-8pyty/dataset/1} \\
3 & Traffic sign & detec & \url{https://universe.roboflow.com/hima-mgeqn/traffic-sign-0yf6k/dataset/5} \\
4 & pedestrian Traffic Light & detec & \url{https://universe.roboflow.com/ono-gedd7/pedestrian-traffic-light-puf4a/dataset/4} \\
5 & Pedestrian Traffic Lights detection & classify & \url{https://data.mendeley.com/datasets/9tm59d3nsn/1} \\
6 & pedestrian traffic light detection & detec & \url{https://www.kaggle.com/datasets/yesangwoo/pedestrian-traffic-light-detection-dataset?select=ped_traffic} \\
7 & Pedestrian Lights & detec & \url{https://universe.roboflow.com/test-xi95g/pedestrian-lights-ha1zc/dataset/2} \\
8 & Crosswalk\&pedestrian light & detec & \url{https://universe.roboflow.com/cap-8nhra/crosswalk-pedestrian-light/browse?queryText=&pageSize=50&startingIndex=0&browseQuery=true} \\
9 & traffic signs & detec & \url{https://universe.roboflow.com/jy-ij8pd/traffic-signs-isqqy/dataset/6} \\
10 & dataset & detec & \url{https://universe.roboflow.com/inc-hymh5/dataset-jq3zi/dataset/5} \\
11 & FYP pedestrian traffic lights & detec & \url{https://universe.roboflow.com/fyp-wnch0/fyp-pedestrian-traffic-lights/dataset/10} \\
12 & pedestrian lights & detec & \url{https://universe.roboflow.com/fyp-crossing-roads/pedestrian-lights-0bbig} \\
13 & pedestrian light, crosswalk & detec & \url{https://universe.roboflow.com/chanyoung/pedestrian-light-crosswalk/dataset/2} \\
\bottomrule
\end{tabularx}
\end{center}
\vspace{-2mm}
\begin{itemize}
    \item \textbf{Source platforms:} Row~5 is hosted on Mendeley Data; Row~6 is hosted on Kaggle; all other repositories are hosted on Roboflow Universe.
    \item \textbf{Curation notes:} Row~5 contains raw images for binary red/green classification; Rows~1, 2, and~4 require class-name standardization or relabeling; Row~3 contains pedestrian crossing warning signs rather than signal heads; Row~8 also contains pedestrian pushbuttons; and Row~10 mixes vehicular stop signs that must be excluded.
\end{itemize}

\subsection*{Accessible Pedestrian Signal (APS) / Push Button}
\vspace{-2mm}
\begin{center}
\scriptsize
\setlength{\tabcolsep}{4pt}
\renewcommand{\arraystretch}{1.04}
\captionof{table}{Candidate public repositories surveyed for \texttt{aps\_button}.}
\label{tab:curation-aps-button}
\begin{tabularx}{\textwidth}{r p{3.6cm} l X}
\toprule
\textbf{\#} & \textbf{Project Name} & \textbf{Type} & \textbf{URL} \\
\midrule
1 & ped-push-button & detec & \url{https://universe.roboflow.com/garbage-y0auo/ped-push-button/dataset/1} \\
2 & ped-push-buttonv1 & detec & \url{https://universe.roboflow.com/akash-lozqg/ped-push-buttonv1/dataset/1} \\
3 & completeData Old+New & detec & \url{https://universe.roboflow.com/akash-lozqg/completedata-old-new/dataset/1} \\
4 & assfasfsa & detec & \url{https://universe.roboflow.com/akash-lozqg/assfasfsa/dataset/1} \\
\bottomrule
\end{tabularx}
\end{center}
\vspace{-2mm}
\begin{itemize}
    \item \textbf{Source platform:} All surveyed repositories are hosted on Roboflow Universe.
    \item \textbf{Curation notes:} All four repositories provide bounding-box detections rather than segmentation masks (with Rows~1 and~2 containing cluttered multi-object scenes and Row~4 duplicating Row~3), leaving zero usable open-source segmentation masks for \texttt{aps\_button}.
\end{itemize}

\clearpage
\subsection*{Bus Stop}
\vspace{-2mm}
\begin{center}
\scriptsize
\setlength{\tabcolsep}{4pt}
\renewcommand{\arraystretch}{1.04}
\captionof{table}{Candidate public repositories surveyed for \texttt{bus\_stop}.}
\label{tab:curation-bus-stop}
\begin{tabularx}{\textwidth}{r p{3.6cm} l X}
\toprule
\textbf{\#} & \textbf{Project Name} & \textbf{Type} & \textbf{URL} \\
\midrule
1 & bus stop identification & detec & \url{https://universe.roboflow.com/project/bus-stop-identification/dataset/1/images?split=train} \\
2 & GQP-Bus-Stops & detec & \url{https://universe.roboflow.com/abhishek-shivdeo/gqp-bus-stops/dataset/1} \\
3 & Traffic & detec & \url{https://universe.roboflow.com/ricky-z8hgp/traffic-v6j6h/browse?queryText=class\%3A\%22bus+stop+number+plate\%22&pageSize=50&startingIndex=0&browseQuery=true} \\
4 & Busstop detection & detec & \url{https://universe.roboflow.com/bus-stop-anatomy/busstop-detection/dataset/5} \\
5 & Bus Stop Object Detection Retry & detec & \url{https://universe.roboflow.com/bus-stop-anatomy/bus-stop-object-detection-retry-cdbss/browse?queryText=&pageSize=50&startingIndex=0&browseQuery=true} \\
6 & Bus Stop & detec & \url{https://universe.roboflow.com/mtsb/bus-stop-7gy2a/dataset/3/images?split=train} \\
\bottomrule
\end{tabularx}
\end{center}
\vspace{-2mm}
\begin{itemize}
    \item \textbf{Source platform:} All surveyed repositories are hosted on Roboflow Universe.
    \item \textbf{Curation notes:} All six repositories are bounding-box detection sets with severe taxonomy or quality issues (Row~1 contains rotated images; Rows~4 and~5 label sub-parts such as \texttt{bench} and \texttt{shelter} without a unified \texttt{bus\_stop} mask; Row~6 depicts miscellaneous objects near bus stops), resulting in zero retained open-source segmentation masks.
\end{itemize}

\subsection*{Bus Stop Sign}
\vspace{-2mm}
\begin{center}
\scriptsize
\setlength{\tabcolsep}{4pt}
\renewcommand{\arraystretch}{1.04}
\captionof{table}{Candidate public repositories surveyed for \texttt{bus\_stop\_sign}.}
\label{tab:curation-bus-stop-sign}
\begin{tabularx}{\textwidth}{r p{3.6cm} l X}
\toprule
\textbf{\#} & \textbf{Project Name} & \textbf{Type} & \textbf{URL} \\
\midrule
1 & Bus Stop Detection & detec & \url{https://universe.roboflow.com/the-university-of-hong-kong-zsdbj/bus-stop-detection/dataset/4} \\
2 & Bus stop & detec & \url{https://universe.roboflow.com/university-of-florida-ae5pr/bus-stop-afloz/dataset/1} \\
3 & bus stop case & detec & \url{https://universe.roboflow.com/summer-2025/bus-stop-case/dataset/1} \\
4 & Traffic & detec & \url{https://universe.roboflow.com/simon-mkr7w/traffic-nnnm9/browse?queryText=class\%3A\%22bus+stop\%22&pageSize=200&startingIndex=400&browseQuery=true} \\
5 & Bus Stop Detection 2 & detec & \url{https://universe.roboflow.com/summer-2025/bus-stop-detection-2-kl5kq} \\
\bottomrule
\end{tabularx}
\end{center}
\vspace{-2mm}
\begin{itemize}
    \item \textbf{Source platform:} All surveyed repositories are hosted on Roboflow Universe.
    \item \textbf{Curation notes:} All five repositories provide bounding-box detections rather than segmentation masks (with Row~1 containing 4-image collages and Rows~3--5 conflating bus-stop shelters and sign stands under a single label), resulting in zero retained open-source segmentation masks.
\end{itemize}

\clearpage
\input{checklist.tex}

\end{document}

%% file: checklist.tex
\section*{NeurIPS Paper Checklist}

\begin{enumerate}

\item {\bf Claims}
    \item[] Question: Do the main claims made in the abstract and introduction accurately reflect the paper's contributions and scope?
    \item[] Answer: \answerYes{}
    \item[] Justification: The abstract and introduction explicitly state our two technical contributions (the \textsc{NavAble} Dataset and \textsc{NavAble} Generator) and four empirical findings, all of which are substantiated by the experiments in Section~\ref{sec:experiments} and the dataset/pipeline described in Sections ~\ref{sec:navable_pipeline}.

    \item[] Guidelines:
    \begin{itemize}
        \item The answer \answerNA{} means that the abstract and introduction do not include the claims made in the paper.
        \item The abstract and/or introduction should clearly state the claims made, including the contributions made in the paper and important assumptions and limitations. A \answerNo{} or \answerNA{} answer to this question will not be perceived well by the reviewers. 
        \item The claims made should match theoretical and experimental results, and reflect how much the results can be expected to generalize to other settings. 
        \item It is fine to include aspirational goals as motivation as long as it is clear that these goals are not attained by the paper. 
    \end{itemize}

\item {\bf Limitations}
    \item[] Question: Does the paper discuss the limitations of the work performed by the authors?
    \item[] Answer: \answerYes{}
    \item[] Justification: Limitations are discussed throughout the paper, including segmentation-quality bottlenecks for long-tail classes in Section~\ref{sec:navable_pipeline}, and the geographic scope of our collected real-world data in Section~\ref{collected-data}. We also discuss limitations in the Section ~\ref{sec:conclusion}

    \item[] Guidelines:
    \begin{itemize}
        \item The answer \answerNA{} means that the paper has no limitation while the answer \answerNo{} means that the paper has limitations, but those are not discussed in the paper. 
        \item The authors are encouraged to create a separate ``Limitations'' section in their paper.
        \item The paper should point out any strong assumptions and how robust the results are to violations of these assumptions (e.g., independence assumptions, noiseless settings, model well-specification, asymptotic approximations only holding locally). The authors should reflect on how these assumptions might be violated in practice and what the implications would be.
        \item The authors should reflect on the scope of the claims made, e.g., if the approach was only tested on a few datasets or with a few runs. In general, empirical results often depend on implicit assumptions, which should be articulated.
        \item The authors should reflect on the factors that influence the performance of the approach. For example, a facial recognition algorithm may perform poorly when image resolution is low or images are taken in low lighting. Or a speech-to-text system might not be used reliably to provide closed captions for online lectures because it fails to handle technical jargon.
        \item The authors should discuss the computational efficiency of the proposed algorithms and how they scale with dataset size.
        \item If applicable, the authors should discuss possible limitations of their approach to address problems of privacy and fairness.
        \item While the authors might fear that complete honesty about limitations might be used by reviewers as grounds for rejection, a worse outcome might be that reviewers discover limitations that aren't acknowledged in the paper. The authors should use their best judgment and recognize that individual actions in favor of transparency play an important role in developing norms that preserve the integrity of the community. Reviewers will be specifically instructed to not penalize honesty concerning limitations.
    \end{itemize}

\item {\bf Theory assumptions and proofs}
    \item[] Question: For each theoretical result, does the paper provide the full set of assumptions and a complete (and correct) proof?
    \item[] Answer: \answerNA{} 
    \item[] Justification: This is an empirical dataset and pipeline paper and does not include theoretical results.
    \item[] Guidelines:
    \begin{itemize}
        \item The answer \answerNA{} means that the paper does not include theoretical results. 
        \item All the theorems, formulas, and proofs in the paper should be numbered and cross-referenced.
        \item All assumptions should be clearly stated or referenced in the statement of any theorems.
        \item The proofs can either appear in the main paper or the supplemental material, but if they appear in the supplemental material, the authors are encouraged to provide a short proof sketch to provide intuition. 
        \item Inversely, any informal proof provided in the core of the paper should be complemented by formal proofs provided in appendix or supplemental material.
        \item Theorems and Lemmas that the proof relies upon should be properly referenced. 
    \end{itemize}

    \item {\bf Experimental result reproducibility}
    \item[] Question: Does the paper fully disclose all the information needed to reproduce the main experimental results of the paper to the extent that it affects the main claims and/or conclusions of the paper (regardless of whether the code and data are provided or not)?
    \item[] Answer: \answerYes{} 
    \item[] Justification: We release the full dataset, synthetic data generation pipeline, and Isaac~Sim extension at \url{https://huggingface.co/datasets/NavAble/NeurIPS_2026_BLV}, with training and evaluation details documented in Section~\ref{subsec:setup} and Appendix~\ref{app:training_details}.
    \item[] Guidelines:
    \begin{itemize}
        \item The answer \answerNA{} means that the paper does not include experiments.
        \item If the paper includes experiments, a \answerNo{} answer to this question will not be perceived well by the reviewers: Making the paper reproducible is important, regardless of whether the code and data are provided or not.
        \item If the contribution is a dataset and\slash or model, the authors should describe the steps taken to make their results reproducible or verifiable. 
        \item Depending on the contribution, reproducibility can be accomplished in various ways. For example, if the contribution is a novel architecture, describing the architecture fully might suffice, or if the contribution is a specific model and empirical evaluation, it may be necessary to either make it possible for others to replicate the model with the same dataset, or provide access to the model. In general. releasing code and data is often one good way to accomplish this, but reproducibility can also be provided via detailed instructions for how to replicate the results, access to a hosted model (e.g., in the case of a large language model), releasing of a model checkpoint, or other means that are appropriate to the research performed.
        \item While NeurIPS does not require releasing code, the conference does require all submissions to provide some reasonable avenue for reproducibility, which may depend on the nature of the contribution. For example
        \begin{enumerate}
            \item If the contribution is primarily a new algorithm, the paper should make it clear how to reproduce that algorithm.
            \item If the contribution is primarily a new model architecture, the paper should describe the architecture clearly and fully.
            \item If the contribution is a new model (e.g., a large language model), then there should either be a way to access this model for reproducing the results or a way to reproduce the model (e.g., with an open-source dataset or instructions for how to construct the dataset).
            \item We recognize that reproducibility may be tricky in some cases, in which case authors are welcome to describe the particular way they provide for reproducibility. In the case of closed-source models, it may be that access to the model is limited in some way (e.g., to registered users), but it should be possible for other researchers to have some path to reproducing or verifying the results.
        \end{enumerate}
    \end{itemize}

\item {\bf Open access to data and code}
    \item[] Question: Does the paper provide open access to the data and code, with sufficient instructions to faithfully reproduce the main experimental results, as described in supplemental material?
    \item[] Answer: \answerYes{} 
    \item[] Justification: The dataset and code are publicly available at the project repository (\url{[https://guidedogrobot-navable.github.io](https://guidedogrobot-navable.github.io)}), with a README providing instructions for dataset usage, model fine-tuning, and running the synthetic data generation pipeline.
    \item[] Guidelines:
    \begin{itemize}
        \item The answer \answerNA{} means that paper does not include experiments requiring code.
        \item Please see the NeurIPS code and data submission guidelines (\url{https://neurips.cc/public/guides/CodeSubmissionPolicy}) for more details.
        \item While we encourage the release of code and data, we understand that this might not be possible, so \answerNo{} is an acceptable answer. Papers cannot be rejected simply for not including code, unless this is central to the contribution (e.g., for a new open-source benchmark).
        \item The instructions should contain the exact command and environment needed to run to reproduce the results. See the NeurIPS code and data submission guidelines (\url{https://neurips.cc/public/guides/CodeSubmissionPolicy}) for more details.
        \item The authors should provide instructions on data access and preparation, including how to access the raw data, preprocessed data, intermediate data, and generated data, etc.
        \item The authors should provide scripts to reproduce all experimental results for the new proposed method and baselines. If only a subset of experiments are reproducible, they should state which ones are omitted from the script and why.
        \item At submission time, to preserve anonymity, the authors should release anonymized versions (if applicable).
        \item Providing as much information as possible in supplemental material (appended to the paper) is recommended, but including URLs to data and code is permitted.
    \end{itemize}

\item {\bf Experimental setting/details}
    \item[] Question: Does the paper specify all the training and test details (e.g., data splits, hyperparameters, how they were chosen, type of optimizer) necessary to understand the results?
    \item[] Answer: \answerYes{} 
    \item[] Justification: Architectures, training-data configurations, augmentations, optimizers, learning-rate schedules, and per-configuration iteration budgets are specified in Section~\ref{subsec:setup} and Appendix~\ref{app:training_details}, with checkpoints selected by validation mIoU on a held-out real split.
    \item[] Guidelines:
    \begin{itemize}
        \item The answer \answerNA{} means that the paper does not include experiments.
        \item The experimental setting should be presented in the core of the paper to a level of detail that is necessary to appreciate the results and make sense of them.
        \item The full details can be provided either with the code, in appendix, or as supplemental material.
    \end{itemize}

\item {\bf Experiment statistical significance}
    \item[] Question: Does the paper report error bars suitably and correctly defined or other appropriate information about the statistical significance of the experiments?
    \item[] Answer: \answerNo{} 
    \item[] Justification: Error bars across multiple seeds are not reported because each fine-tuning run is computationally expensive (multi-GPU training over 16K--324K iterations); all configurations share identical validation/test splits and checkpoint-selection protocols to ensure fair comparison.
    \item[] Guidelines:
    \begin{itemize}
        \item The answer \answerNA{} means that the paper does not include experiments.
        \item The authors should answer \answerYes{} if the results are accompanied by error bars, confidence intervals, or statistical significance tests, at least for the experiments that support the main claims of the paper.
        \item The factors of variability that the error bars are capturing should be clearly stated (for example, train/test split, initialization, random drawing of some parameter, or overall run with given experimental conditions).
        \item The method for calculating the error bars should be explained (closed form formula, call to a library function, bootstrap, etc.)
        \item The assumptions made should be given (e.g., Normally distributed errors).
        \item It should be clear whether the error bar is the standard deviation or the standard error of the mean.
        \item It is OK to report 1-sigma error bars, but one should state it. The authors should preferably report a 2-sigma error bar than state that they have a 96\% CI, if the hypothesis of Normality of errors is not verified.
        \item For asymmetric distributions, the authors should be careful not to show in tables or figures symmetric error bars that would yield results that are out of range (e.g., negative error rates).
        \item If error bars are reported in tables or plots, the authors should explain in the text how they were calculated and reference the corresponding figures or tables in the text.
    \end{itemize}

\item {\bf Experiments compute resources}
    \item[] Question: For each experiment, does the paper provide sufficient information on the computer resources (type of compute workers, memory, time of execution) needed to reproduce the experiments?
    \item[] Answer: \answerYes{} 
    \item[] Justification: Appendix~\ref{app:training_details} specifies the training hardware (NVIDIA L40S 48\,GB cluster nodes and RTX PRO 6000 Blackwell Max-Q workstations) along with per-architecture, per-configuration iteration budgets.
    \item[] Guidelines:
    \begin{itemize}
        \item The answer \answerNA{} means that the paper does not include experiments.
        \item The paper should indicate the type of compute workers CPU or GPU, internal cluster, or cloud provider, including relevant memory and storage.
        \item The paper should provide the amount of compute required for each of the individual experimental runs as well as estimate the total compute. 
        \item The paper should disclose whether the full research project required more compute than the experiments reported in the paper (e.g., preliminary or failed experiments that didn't make it into the paper). 
    \end{itemize}
    
\item {\bf Code of ethics}
    \item[] Question: Does the research conducted in the paper conform, in every respect, with the NeurIPS Code of Ethics \url{https://neurips.cc/public/EthicsGuidelines}?
    \item[] Answer: \answerYes{} 
    \item[] Justification: The research conforms to the NeurIPS Code of Ethics; data was collected in publicly accessible environments, no human subjects were involved, and personally identifiable information was carefully removed from our collected data.
    \item[] Guidelines:
    \begin{itemize}
        \item The answer \answerNA{} means that the authors have not reviewed the NeurIPS Code of Ethics.
        \item If the authors answer \answerNo, they should explain the special circumstances that require a deviation from the Code of Ethics.
        \item The authors should make sure to preserve anonymity (e.g., if there is a special consideration due to laws or regulations in their jurisdiction).
    \end{itemize}

\item {\bf Broader impacts}
    \item[] Question: Does the paper discuss both potential positive societal impacts and negative societal impacts of the work performed?
    \item[] Answer: \answerYes{} 
    \item[] Justification: The abstract and introduction motivate the work by its positive societal impact on assistive perception for the estimated 2.2 billion people with vision impairment, enabling safer and more independent navigation for BLV individuals.
    \item[] Guidelines:
    \begin{itemize}
        \item The answer \answerNA{} means that there is no societal impact of the work performed.
        \item If the authors answer \answerNA{} or \answerNo, they should explain why their work has no societal impact or why the paper does not address societal impact.
        \item Examples of negative societal impacts include potential malicious or unintended uses (e.g., disinformation, generating fake profiles, surveillance), fairness considerations (e.g., deployment of technologies that could make decisions that unfairly impact specific groups), privacy considerations, and security considerations.
        \item The conference expects that many papers will be foundational research and not tied to particular applications, let alone deployments. However, if there is a direct path to any negative applications, the authors should point it out. For example, it is legitimate to point out that an improvement in the quality of generative models could be used to generate Deepfakes for disinformation. On the other hand, it is not needed to point out that a generic algorithm for optimizing neural networks could enable people to train models that generate Deepfakes faster.
        \item The authors should consider possible harms that could arise when the technology is being used as intended and functioning correctly, harms that could arise when the technology is being used as intended but gives incorrect results, and harms following from (intentional or unintentional) misuse of the technology.
        \item If there are negative societal impacts, the authors could also discuss possible mitigation strategies (e.g., gated release of models, providing defenses in addition to attacks, mechanisms for monitoring misuse, mechanisms to monitor how a system learns from feedback over time, improving the efficiency and accessibility of ML).
    \end{itemize}
    
\item {\bf Safeguards}
    \item[] Question: Does the paper describe safeguards that have been put in place for responsible release of data or models that have a high risk for misuse (e.g., pre-trained language models, image generators, or scraped datasets)?
    \item[] Answer: \answerNA{} 
    \item[] Justification: We do not release any pretrained generative models or other artifacts with high misuse risk; the released artifacts are a segmentation dataset, a 3D asset library, and a data generation pipeline.
    \item[] Guidelines:
    \begin{itemize}
        \item The answer \answerNA{} means that the paper poses no such risks.
        \item Released models that have a high risk for misuse or dual-use should be released with necessary safeguards to allow for controlled use of the model, for example by requiring that users adhere to usage guidelines or restrictions to access the model or implementing safety filters. 
        \item Datasets that have been scraped from the Internet could pose safety risks. The authors should describe how they avoided releasing unsafe images.
        \item We recognize that providing effective safeguards is challenging, and many papers do not require this, but we encourage authors to take this into account and make a best faith effort.
    \end{itemize}

\item {\bf Licenses for existing assets}
    \item[] Question: Are the creators or original owners of assets (e.g., code, data, models), used in the paper, properly credited and are the license and terms of use explicitly mentioned and properly respected?
    \item[] Answer: \answerYes{} 
    \item[] Justification: We are the creators of the newly collected real-world data, synthetic assets, and pipeline; we ensure proper licensing for curated open-source data and third-party environments (e.g., Fab marketplace) on our HuggingFace repository.
    \item[] Guidelines:
    \begin{itemize}
        \item The answer \answerNA{} means that the paper does not use existing assets.
        \item The authors should cite the original paper that produced the code package or dataset.
        \item The authors should state which version of the asset is used and, if possible, include a URL.
        \item The name of the license (e.g., CC-BY 4.0) should be included for each asset.
        \item For scraped data from a particular source (e.g., website), the copyright and terms of service of that source should be provided.
        \item If assets are released, the license, copyright information, and terms of use in the package should be provided. For popular datasets, \url{paperswithcode.com/datasets} has curated licenses for some datasets. Their licensing guide can help determine the license of a dataset.
        \item For existing datasets that are re-packaged, both the original license and the license of the derived asset (if it has changed) should be provided.
        \item If this information is not available online, the authors are encouraged to reach out to the asset's creators.
    \end{itemize}

\item {\bf New assets}
    \item[] Question: Are new assets introduced in the paper well documented and is the documentation provided alongside the assets?
    \item[] Answer: \answerYes{} 
    \item[] Justification: Documentation, dataset cards, and usage instructions for the \textsc{NavAble} dataset, 3D asset library, and Isaac~Sim extension are provided alongside the release in the linked repository.
    \item[] Guidelines:
    \begin{itemize}
        \item The answer \answerNA{} means that the paper does not release new assets.
        \item Researchers should communicate the details of the dataset\slash code\slash model as part of their submissions via structured templates. This includes details about training, license, limitations, etc. 
        \item The paper should discuss whether and how consent was obtained from people whose asset is used.
        \item At submission time, remember to anonymize your assets (if applicable). You can either create an anonymized URL or include an anonymized zip file.
    \end{itemize}

\item {\bf Crowdsourcing and research with human subjects}
    \item[] Question: For crowdsourcing experiments and research with human subjects, does the paper include the full text of instructions given to participants and screenshots, if applicable, as well as details about compensation (if any)? 
    \item[] Answer: \answerNA{} 
    \item[] Justification: The paper does not involve crowdsourcing or research with human subjects; all annotations were performed by the authors.
    \item[] Guidelines:
    \begin{itemize}
        \item The answer \answerNA{} means that the paper does not involve crowdsourcing nor research with human subjects.
        \item Including this information in the supplemental material is fine, but if the main contribution of the paper involves human subjects, then as much detail as possible should be included in the main paper. 
        \item According to the NeurIPS Code of Ethics, workers involved in data collection, curation, or other labor should be paid at least the minimum wage in the country of the data collector. 
    \end{itemize}

\item {\bf Institutional review board (IRB) approvals or equivalent for research with human subjects}
    \item[] Question: Does the paper describe potential risks incurred by study participants, whether such risks were disclosed to the subjects, and whether Institutional Review Board (IRB) approvals (or an equivalent approval/review based on the requirements of your country or institution) were obtained?
    \item[] Answer: \answerNA{} 
    \item[] Justification: The paper does not involve research with human subjects.
    \item[] Guidelines:
    \begin{itemize}
        \item The answer \answerNA{} means that the paper does not involve crowdsourcing nor research with human subjects.
        \item Depending on the country in which research is conducted, IRB approval (or equivalent) may be required for any human subjects research. If you obtained IRB approval, you should clearly state this in the paper. 
        \item We recognize that the procedures for this may vary significantly between institutions and locations, and we expect authors to adhere to the NeurIPS Code of Ethics and the guidelines for their institution. 
        \item For initial submissions, do not include any information that would break anonymity (if applicable), such as the institution conducting the review.
    \end{itemize}

\item {\bf Declaration of LLM usage}
    \item[] Question: Does the paper describe the usage of LLMs if it is an important, original, or non-standard component of the core methods in this research? Note that if the LLM is used only for writing, editing, or formatting purposes and does \emph{not} impact the core methodology, scientific rigor, or originality of the research, declaration is not required.
    \item[] Answer: \answerYes{} 
    \item[] Justification: We use a vision-language model (Gemini) as a non-trivial component for filtering web-crawled images during synthetic asset generation, as described in Section~\ref{sec:navable_pipeline}.
    \item[] Guidelines:
    \begin{itemize}
        \item The answer \answerNA{} means that the core method development in this research does not involve LLMs as any important, original, or non-standard components.
        \item Please refer to our LLM policy in the NeurIPS handbook for what should or should not be described.
    \end{itemize}

\end{enumerate}